\documentclass{article} %
\usepackage{iclr2026_conference,times}
\iclrfinalcopy

\usepackage{amsmath,amsfonts,bm}

\def\eqref#1{equation~\ref{#1}}

\def\1{\bm{1}}

\DeclareMathAlphabet{\mathsfit}{\encodingdefault}{\sfdefault}{m}{sl}
\SetMathAlphabet{\mathsfit}{bold}{\encodingdefault}{\sfdefault}{bx}{n}

\usepackage[utf8]{inputenc} %
\usepackage[T1]{fontenc}    %
\usepackage{hyperref}       %
\usepackage{url}            %
\usepackage{booktabs}       %
\usepackage{amsfonts}       %
\usepackage{nicefrac}       %
\usepackage{microtype}      %
\usepackage{xcolor}         %

\usepackage{adjustbox}        %
\usepackage{algorithm}        %
\usepackage{algpseudocode}    %
\usepackage{amsfonts}         %
\usepackage{amsmath}          %
\usepackage{amssymb}          %
\usepackage{bm}               %
\usepackage{booktabs}         %
\usepackage{colortbl}         %
\usepackage{comment}          %
\usepackage{dsfont}           %
\usepackage{enumitem}         %
\usepackage{hyperref}         %
\usepackage{listings}         %
\usepackage{lscape}           %
\usepackage{makecell}         %
\usepackage{mathtools}        %
\usepackage{multirow}         %
\usepackage{multicol}         %
\usepackage{nicefrac}         %
\usepackage{pifont}           %
\usepackage{scalerel}         %
\usepackage{soul}             %
\usepackage{subcaption}       %
\usepackage{tabularx}         %
\usepackage{tablefootnote}    %
\usepackage{tcolorbox}        %
\usepackage{url}              %
\usepackage{wrapfig}          %
\usepackage{xcolor}           %
\usepackage{xspace}           %
\usepackage{tabularray}

\lstdefinestyle{pythonstyle}{
    language=Python,
    basicstyle=\ttfamily\small,
    keywordstyle=\color{blue}\bfseries,
    stringstyle=\color{orange},
    commentstyle=\color{blue},
    showstringspaces=false,
    frame=single,
    breaklines=true
}

\title{\centering \titlesealicon~\bsc{SealQA}: Raising the Bar for Reasoning in \\ Search-Augmented Language Models}

\newcommand{\myhspace}{\hspace{5.0ex}}

\author{
\centerline{Thinh Pham \myhspace Nguyen Nguyen \myhspace Pratibha Zunjare} \\[1.0ex] \bf \centerline{Weiyuan Chen \myhspace Yu-Min Tseng \myhspace Tu Vu} \\ [1.0ex]
\centerline{Virginia Tech} \\ [1.0ex]
\centerline{\texttt{\{thinhphp,tuvu\}@vt.edu}}
}
\vspace{-2mm}

\newcommand{\titlesealicon}[0]{\includegraphics[height=.03\textheight,width=.055\textwidth]{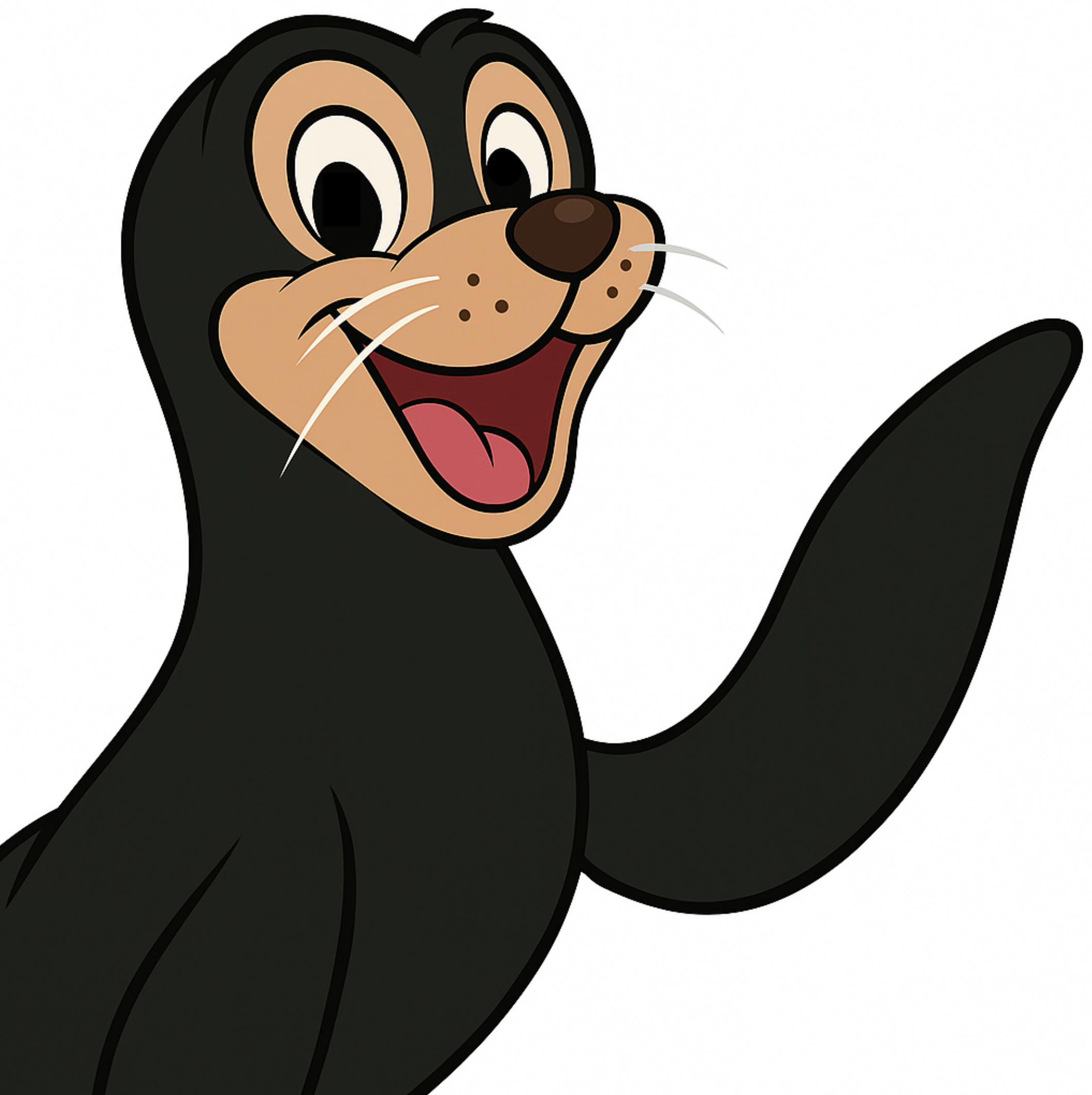}}
\newcommand{\ssc}[1]{{\small \sc #1}\xspace}
\newcommand{\sscnoxsp}[1]{{\small \sc #1}}
\newcommand{\bsc}[1]{{\sc \textbf{#1}}\xspace}
\newcommand{\bssc}[1]{{\footnotesize \sc \textbf{#1}}\xspace}

\newcommand{\fssc}[1]{{\scriptsize \sc #1}\xspace}

\newcommand{\sealqa}{\ssc{SealQA}}
\newcommand{\sealzero}{\ssc{Seal-0}}
\newcommand{\sealhard}{\ssc{Seal-Hard}}
\newcommand{\longseal}{\ssc{LongSeal}}

\newcommand{\google}{\ssc{Google}}
\newcommand{\urls}{\ssc{URLs}}
\newcommand{\qa}{\ssc{QA}}

\newcommand{\nlp}{\ssc{NLP}}

\newcommand{\llms}{\ssc{LLMs}}
\newcommand{\api}{\ssc{API}}

\newcommand{\freshprompt}{\ssc{FreshPrompt}}
\newcommand{\selfask}{\ssc{self-ask}}

\newcommand{\chat}{\ssc{Chat}}
\newcommand{\reason}{\ssc{Reason.}}
\newcommand{\agentic}{\ssc{Agentic}}
\newcommand{\deepseek}{\ssc{DeepSeek-R1-671B}}
\newcommand{\newdeepseek}{\ssc{DeepSeek-R1-0528-671B}}

\newcommand{\smallurl}[1]{ \begin{scriptsize}\url{#1}\end{scriptsize}}

\definecolor{myblue}{HTML}{2B79B0}
\definecolor{myorange}{HTML}{FB7F36}
\definecolor{linkcolor}{HTML}{389E3B}
\definecolor{mygreen}{HTML}{DFFFDF}
\definecolor{myred}{HTML}{D22D35}
\definecolor{myyellow}{HTML}{FFF2A6}
\definecolor{mygolden}{HTML}{FDD83B}

\date{}

\begin{document}
\maketitle
\vspace{-4mm}
\begin{abstract}
We introduce \bssc{SealQA}, a challenge benchmark for evaluating \textbf{SE}arch-\textbf{A}ugmented \textbf{L}anguage models on fact-seeking questions where web search yields conflicting, noisy, or unhelpful results. %
\sealqa comes in \emph{three} flavors: (1) \bssc{Seal-0} (\emph{main}) and (2) \bssc{Seal-Hard}, both of which assess factual accuracy and reasoning capabilities, where \sealzero %
targets the most challenging questions that frontier non-reasoning models (e.g., \ssc{GPT-4.1}) answer with near-zero accuracy%
; and (3) \bssc{LongSeal}, which extends \sealqa to test long-context, multi-document reasoning in \emph{``needle-in-a-haystack''} settings. Our evaluation reveals critical limitations in current models. Even frontier reasoning models face significant challenges across \sealqa flavors. On \sealzero, \ssc{GPT-5} with tools achieves only 43.2\% accuracy at its best reasoning effort. We also find that even advanced reasoning models (e.g., \ssc{DeepSeek-R1}) can be vulnerable to noisy search results. \emph{Notably}, increasing test-time compute does not yield reliable gains across \ssc{GPT-5} and the \ssc{o}-series of models, with performance often plateauing or even declining early. Finally, while current models are less affected by the \emph{``lost-in-the-middle''} issue, they still fail to reliably identify relevant documents in \longseal when faced with numerous distractors. %
To facilitate future work, we release \sealqa at \href{https://huggingface.co/datasets/vtllms/sealqa}{\textcolor{linkcolor}{\texttt{https://huggingface.co/datasets/vtllms/sealqa}}}.
\end{abstract}
\vspace{-4mm}

\section{Introduction}
\vspace{-5mm}
\begin{figure}[h!]
\centering
\includegraphics[width=\textwidth]{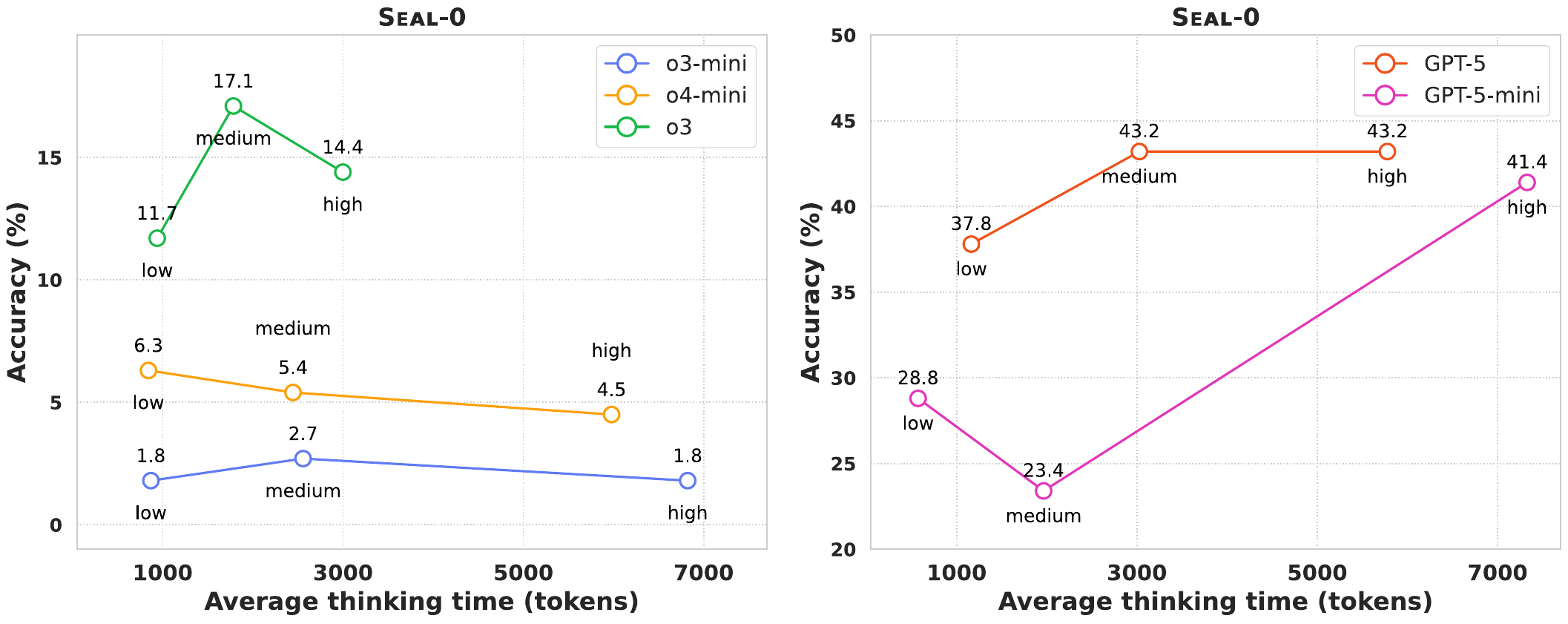}
\caption{Frontier model performance before (left) and after (right) the release of \sealzero. Despite potential data contamination or direct access by agentic models, \sealzero continues to pose a significant challenge for current frontier LLMs. Test-time scaling does not lead to reliable gains, with performance often plateauing or even declining early.  See Figure~\ref{figure:test_time_scaling_sealhard} and Table~\ref{table:additional_seal0_results} for additional results.
}
\label{figure:test_time_scaling_seal0}
\vspace{-2mm}
\end{figure}

Large language models (\llms) have entered a new scaling paradigm: \emph{test-time scaling}, where models dynamically allocate more compute during inference time to improve performance~\citep{openai2025gpt5,google2025gemini,xai2025grok4,anthropic2025introducing}.
This paradigm shift is embodied in \emph{reasoning models}, which leverage reinforcement learning and other techniques to guide inference-time strategies such as chain-of-thought reasoning, recursive refinement, and real-time search~\citep{muennighoff2025s1,guo2025deepseek,snell2024scaling,geiping2025scaling}. These models can now decompose questions into subqueries, decide when and how to query a search engine, %
and fuse retrieved content into structured reasoning paths~\citep{openai2025gpt5,google2025gemini,jin2025search}.
\vspace{-2mm}

As \llms advance, benchmarks that rely on static knowledge and simple reasoning become saturated and fail to keep pace. For example, frontier models now achieve over 90\% accuracy on \ssc{MMLU}~\citep{phan2025humanity}.
Furthermore, most evaluations of search-augmented LLMs focus on short factual queries that top-ranked results answer directly~\citep{vu-etal-2024-freshllms,NEURIPS2023_9941624e}. These setups require only shallow comprehension and fail to reflect the messy, ambiguous nature of real-world search.

\vspace{-1mm}
To properly evaluate today's \llms, benchmarks that go beyond simple fact lookup are needed. Real-world search often returns documents that are outdated, misleading, or superficially relevant but ultimately unhelpful. Navigating this noise requires deeper reasoning that filters inconsistencies, reconciles contradictions, and identifies trustworthy signals. Benchmarks that simulate these challenges are rare, partly because they are difficult to curate and validate at scale.

\begin{figure*}[t]
\vspace{-1.5em}
    \centering
\includegraphics[width=0.8\textwidth]{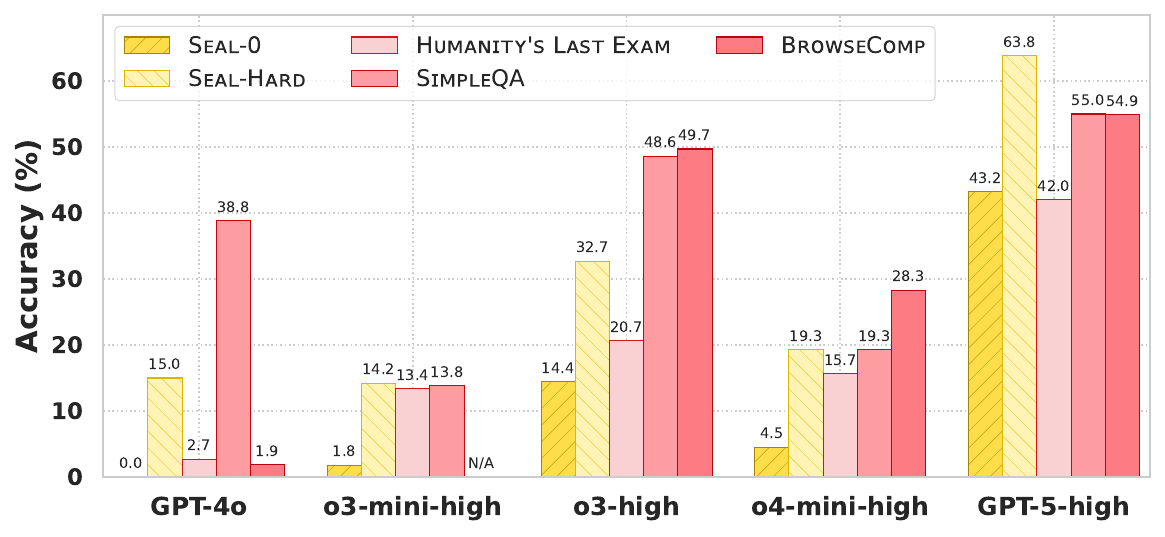}
\caption{Accuracy of \sscnoxsp{LLMs } across benchmarks. \sscnoxsp{SealQA } poses significant challenges to frontier models.}
\label{figure:sealqa_and_friends_fig}
\vspace{-0.8em}
\end{figure*}

\begin{figure*}[t]
\centering
\includegraphics[width=0.95\textwidth]{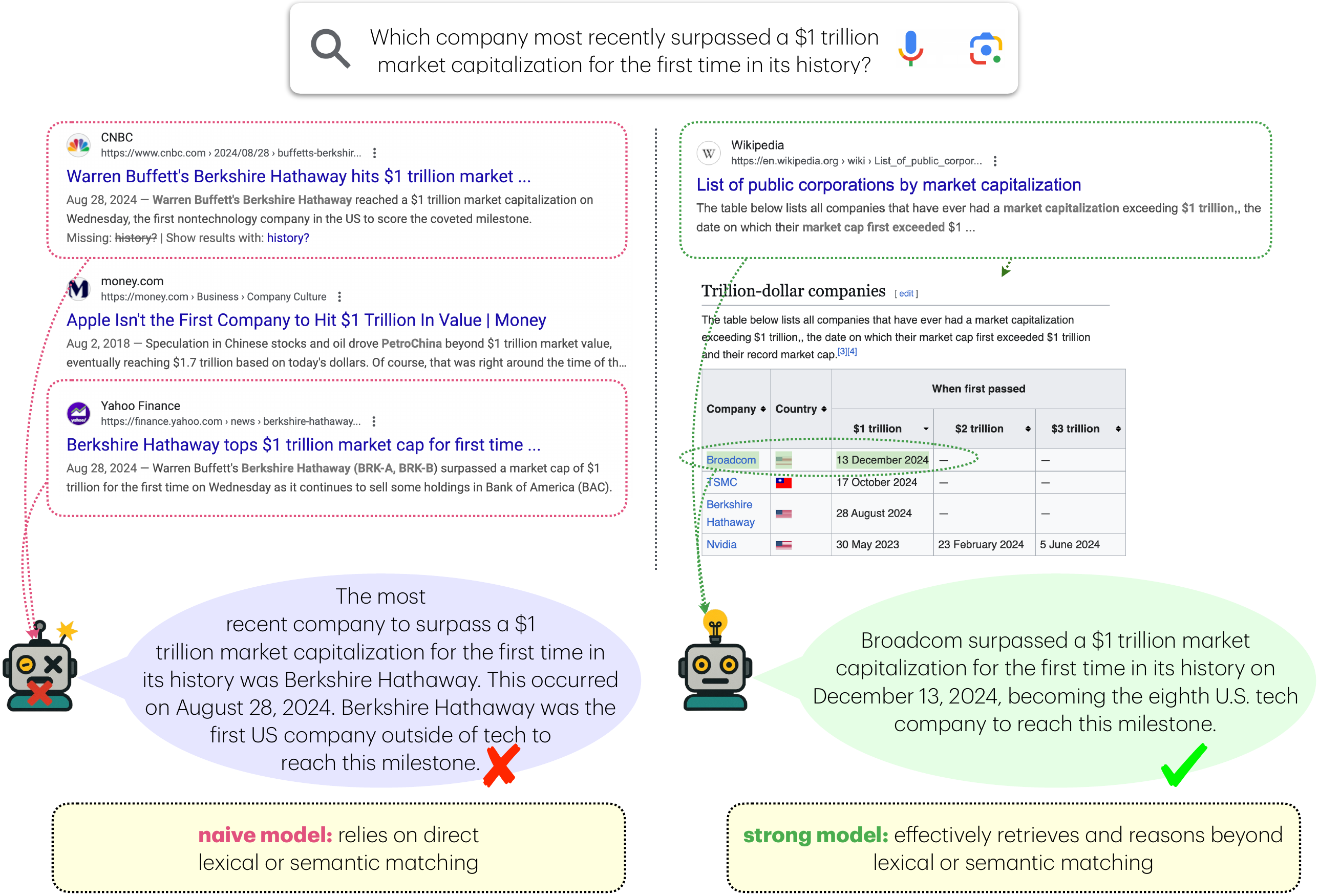}
\caption{\sscnoxsp{SealQA} requires intensive reasoning to resolve ambiguity, filter out misinformation, or reconcile conflicting evidence. See Appendix~\ref{section:appendix_sample_model_outputs} for sample model outputs.}
\label{figure:noisy_search_fig}
\vspace{-5mm}
\end{figure*}

\vspace{-1mm}
We introduce \sealqa , a \emph{small} but \emph{extremely challenging} benchmark (see Figure~\ref{figure:sealqa_and_friends_fig}) for evaluating search-augmented \llms on fact-seeking questions. %
Each \sealqa question is carefully crafted by \nlp researchers to trigger ambiguous, conflicting, or noisy search results (see Figure~\ref{figure:noisy_search_fig}). This design makes it difficult to answer questions through simple keyword matching or by relying on top-ranked documents. \sealqa spans a range of question types, including time-sensitive questions, across diverse domains such as science, technology, sports, entertainment, politics, history, geography, etc.

\sealqa questions probe a broad spectrum of complex reasoning skills. These include distinguishing between similar entities or events, tracking changes to the same entity over time, interpreting information embedded in search-result plots, charts, or tables, counting multiple items, reasoning over non-English content, and debunking false premises or common misconceptions (see Figure~\ref{figure:sealqa_sample_examples_fig} in Appendix~\ref{section:appendix_sample_questions} for sample questions). All questions are self-contained, verifiable, and require intensive reasoning to resolve ambiguity, filter misinformation, or reconcile conflicting evidence. These capabilities are central to modern \llms but are not adequately captured by existing benchmarks. %

To ensure both difficulty and quality, each \sealqa question undergoes a \textit{rigorous} multi-round vetting process: an initial phase with two or more graduate-level reviewers, followed by expert approval.

\vspace{-2mm}
\sealqa comes in \textit{three} \textit{flavors}:
\begin{itemize}[]
\vspace{-1mm}
    \item \sealzero (\emph{main}; 111 questions): A carefully curated core set where frontier non-reasoning models like \ssc{GPT-4.1} with browsing consistently fail. Each question is iteratively refined until multiple models fail across several attempts (0\% accuracy, hence the ``0'' in the name). %
    \vspace{-0.8mm}
    \item \sealhard (254 questions): A broader set that includes \sealzero and additional difficult questions that did not meet our strict failure threshold but remain highly challenging.
    \vspace{-0.8mm}
    \item \longseal (254 questions): A \emph{``needle-in-a-haystack''} variant that tests long-context, multi-document reasoning. Each question is paired with a large set of retrieved documents, among which only one contains or implies the correct answer. This document is buried within irrelevant, noisy, or misleading content.
\end{itemize}
\vspace{-2mm}

We intentionally kept \sealqa small due to the high cost and complexity of question development.\footnote{Each question required over an hour on average -- roughly 45 minutes to draft, plus additional time for review and revision. Many initial ideas were discarded as they failed to meaningfully challenge frontier \fssc{LLMs}.} Building the full benchmark required a team of \emph{six} \nlp researchers working over \emph{eight} months through multiple development cycles. A smaller benchmark also reduces \api evaluation costs, allows more frequent updates, and aligns 
with recent emphasis on high-quality, targeted evaluations over large, noisy ones~\citep{rein2024gpqa,pmlr-v235-maia-polo24a}.\footnote{For example, the widely used \fssc{GPQA-Diamond}~\citep{rein2024gpqa}, a compact set of 198 expert-vetted questions, demonstrates how a small, carefully curated dataset can effectively assess a model's reasoning ability.} \sealqa is also designed for stable evaluation with \emph{low} run-to-run variance.\footnote{Our questions often lead multiple models to fail across repeated attempts.}

Our key contributions are as follows: (1) We introduce \sealqa, a challenge benchmark designed to evaluate reasoning under noisy, conflicting, and ambiguous search results. \sealqa includes three flavors: \sealzero, \sealhard, and \longseal, each targeting different challenges in search-augmented reasoning; (2) We benchmark a range of \llms and uncover significant limitations in current retrieval-augmented approaches. Even \emph{state-of-the-art} models struggle across \sealqa flavors when faced with conflicting or misleading context. On \sealzero, performance remains low even for agentic models equipped with search tools. We also find that advanced reasoning models can be highly vulnerable to noisy search results. \emph{Notably}, increasing test-time compute does not reliably improve performance across \ssc{OpenAI}'s \ssc{GPT-5} and \ssc{o}-series of models --  performance often plateaus or declines. \longseal further reveals major weaknesses in long-context reasoning: while current frontier \llms are more robust to \emph{``lost-in-the-middle''} effects~\citep{liu-etal-2024-lost}, they still fail to reliably identify and prioritize relevant evidence amid distractors; and (3) We publicly release \sscnoxsp{SealQA} as a dynamic, versioned benchmark, and commit to review and update its answers regularly to ensure that evaluations reflect the most recent knowledge.

\vspace{-2mm}
\section{Data collection}
\label{sec:data}
\vspace{-2mm}
In this section, we describe \sealqa, our benchmark designed to capture the complexity of real-world information-seeking. \sealqa rigorously evaluates a model's reasoning ability, robustness to noisy search results, and capacity to handle dynamic, real-world knowledge.
\vspace{-2mm}
\paragraph{Human annotators:} To build \sealqa, we recruited \nlp researchers\footnote{including the authors and their colleagues} as human annotators who were shown a diverse set of exemplars that illustrated the types of questions we sought to collect. %
\vspace{-2mm}
\paragraph{Question types:}\ Our questions span several categories: ({$\mathcal{Q}_1$}) \emph{advanced reasoning}, which covers multi-hop reasoning, interpreting search-result plots, charts, or tables, and performing counting or calculations; ({$\mathcal{Q}_2$}) \emph{entity/event disambiguation}, which focuses on distinguishing between similar entities or events; ({$\mathcal{Q}_3$}) \emph{temporal tracking}, which requires identifying and differentiating instances of entities over time; ({$\mathcal{Q}_4$}) \emph{cross-lingual reasoning}, where the question is in English but answering it requires retrieving and reasoning over non-English sources; and ({$\mathcal{Q}_5$}) \emph{false-premise questions}, which require debunking false assumptions.%
\vspace{-2mm}
\paragraph{Annotation criteria:}  Annotators were instructed to write questions with a \emph{single}, \emph{unambiguous} answer (e.g., specifying ``on what date'' rather than asking ``when''). Each question must be supported by one or more webpages that justify the reference answer, which ensures \emph{verifiability}. For questions that involve \emph{fresh knowledge}, annotators were required to cite regularly updated sources to support future answer updates. We also classify questions by \emph{freshness}~\citep{vu-etal-2024-freshllms}: \emph{never-changing} (\ssc{never}; answers never change), \emph{slow-changing} (\ssc{slow}; answers change over several years), and \emph{fast-changing} (\ssc{fast}; answers typically change within a year). %
All questions were designed to appear natural while still triggering ambiguous, conflicting, or misleading search results when entered into a search engine like \google. Each question has a predefined annotation that classifies its expected search results as \ssc{conflict.} (mixed correct and misleading answers) or \ssc{unhelpful} (no correct answers). %
Annotators also provided explanations for each answer, including any necessary clarification or subtle reasoning. Finally, each question was refined until it consistently caused multiple models to fail across repeated attempts.
\vspace{-2mm}
\paragraph{Quality control:} We employed a rigorous multi-round review process. Each question was first reviewed by two or more graduate-level annotators, followed by expert approval. We performed several rounds of data cleaning, including verification of supporting \urls, answer correctness, and question clarity. %
Questions whose answers change too frequently were excluded. For each question, we also annotated the effective year (when the answer last changed) and the expected next review date to support future maintenance.
\vspace{-2mm}
\paragraph{Diversity:}\ \sealqa questions vary in length, with an average of 31 tokens and a maximum of 69. \sealqa also spans diverse domains: science and technology (26.8\%), sports (22.0\%), entertainment (21.7\%), politics (9.1\%), history and geography (8.3\%), and others (12.2\%).\footnote{Following~\citet{wei2024measuring}, topic labels were assigned post-hoc using \fssc{GPT-4o mini}.} 
By question category, 72.4\% involve advanced reasoning ({$\mathcal{Q}_1$}), 58.3\% entity/event disambiguation ({$\mathcal{Q}_2$}), 13.7\% temporal tracking ({$\mathcal{Q}_3$}), 5.5\% cross-lingual reasoning ({$\mathcal{Q}_4$}), and 4.3\% false-premise detection ({$\mathcal{Q}_5$}). By freshness, 31.1\% are \emph{never-changing}, 43.7\% \emph{slow-changing}, and 25.2\% \emph{fast-changing}. By effective year, 22.0\% reference 2025 events, 19.3\% 2024, and 58.7\% prior to 2024. %
\vspace{-2mm}
\paragraph{Curation of \bssc{SealQA} flavors:} To curate \sealzero, we tested each question against \ssc{GPT-4o}, \ssc{GPT-4.1}, their \ssc{mini} variants~\citep{openai2024gpt4o,openai2024gpt4omini,openai2025introducing}, and \ssc{Llama-4-Scout}~\citep{meta2025llama4}, both with and without browsing.\footnote{We applied \fssc{FreshPrompt}~\citep{vu-etal-2024-freshllms} to \fssc{Llama-4-Scout}.%
} Only questions whose answers all models failed to produce across 10–15 attempts were retained. %
This follows current practices for constructing challenging benchmarks; for example, \ssc{SimpleQA}~\citep{wei2024measuring} was also adversarially collected against \ssc{GPT-4} responses. \sealzero was then combined with other rejected-but-difficult questions to form \sealhard.%

For \longseal, each \sealhard question is paired with a set of retrieved documents: \emph{one} helpful (gold) document from annotator-provided webpages, and up to 50 hard negatives that appear relevant but are unhelpful.\footnote{To collect hard negatives, we used \fssc{Google} to retrieve the top 10 webpages per question and extracted their main content using \fssc{Trafilatura}~\citep{barbaresi-2021-trafilatura}. To add temporal diversity and potential conflicts, we retrieved 10 more pages restricted to pre-2023 content. We also used \fssc{GPT-4o mini} to generate three semantically related queries per question and collected documents for each. Duplicates were removed, and documents whose length exceeded 10K tokens were excluded.} To ensure difficulty, we used \ssc{GPT-4o mini} to filter out negatives whose content might allow the correct answer to be inferred. The gold document was randomly inserted among the negatives. \longseal contains over 7.6K documents and serves as a testbed for long-context reasoning under noisy retrieval conditions.
\vspace{-2mm}
\paragraph{Evaluation protocol:} Models are evaluated using a \ssc{GPT-4o mini} auto-rater adapted from~\citet{wei2024measuring}, which takes the question, predicted answer, and reference answer as input and labels responses as ``correct'', ``incorrect'', or ``not attempted'' (see Appendix~\ref{section:grader_template} for the full prompt). The evaluation follows a relaxed protocol that checks whether the main answer is factually correct and consistent throughout the response.
\vspace{-2mm}
\paragraph{Auto-rater reliability:} To assess the auto-rater's reliability, two authors independently evaluated 100 answers. Disagreements were resolved through discussion, which produced a unified set of human ratings that agreed with the auto-rater 98\% of the time.

\section{Experiments}
\label{sec:exp}
\begingroup
\renewcommand{\arraystretch}{1.3}
\begin{table*}[t]
\centering
\caption{Accuracy on \sealzero and \sealhard.  Frontier \sscnoxsp{LLMs} face significant challenges on \sscnoxsp{SealQA} questions. $^\dagger$ indicates results using \ssc{ChatGPT}'s built-in search; all other search-based results use \ssc{FreshPrompt}~\citep{vu-etal-2024-freshllms}. $^*$ indicates evaluation conducted after the release of \sealqa.}
    \begin{adjustbox}{max width=\textwidth}
    \begin{tabular}{lllcccc}
    \toprule
    \multirow{2}{*}{\textbf{Model}} & \multirow{2}{*}{\textbf{ \shortstack[c]{knowl. \\ cutoff}}} & \multirow{2}{*}{\textbf{type}} & \multicolumn{2}{c}{\sealzero}                                                                & \multicolumn{2}{c}{\sealhard}                \\ \cline{4-7} 
    
    & &  & w/o \ssc{search} & w/ \ssc{search} & w/o \ssc{search} & w/ \ssc{search} \\ \hline
    \rowcolor{myyellow}\multicolumn{7}{c}{\textit{Closed-source models}}  \\ \hline
     \ssc{GPT-4o-mini} & Sep 30, 2023 & \chat & 0.0 & 0.0$^\dagger$ & 9.1 & 13.4$^\dagger$           \\ 
     \ssc{GPT-4.1-mini} & May 31, 2024 & \chat & 0.0 & 0.0$^\dagger$ & 13.8 & 11.8$^\dagger$         \\ 
     \ssc{GPT-4o} & Sep 30, 2023 & \chat & 0.0 & 0.0$^\dagger$ & 11.8 & 15.0$^\dagger$          \\ 
     \ssc{GPT-4.1} & May 31, 2024 & \chat & 0.0  & 0.0$^\dagger$ & 15.0 & 20.5$^\dagger$         \\ 
     \ssc{o3-mini-high} & Sep 30, 2023 & \reason & 3.6 & 1.8 & 12.6 & 14.2         \\ 
     \sscnoxsp{o4-mini-high} & May 31, 2024 & \agentic & - & 4.5 $^\dagger$ & - & 19.3$^\dagger$          \\ 
     \sscnoxsp{o3-high} & May 31, 2024 & \agentic & - & {\color{black}14.4$^\dagger$} & - & {\color{black}32.7$^\dagger$}         \\
     \ssc{GPT-5-mini-high}$^*$ & May 31, 2024 & \reason & 6.3  & 41.4$^\dagger$ & 16.9 & 60.2$^\dagger$  \\
     \ssc{GPT-5-high}$^*$ & Sep 30, 2024 & \reason & \textbf{15.3}  & \textbf{43.2}$^\dagger$ & \textbf{37.8} & \textbf{63.8}$^\dagger$  \\ \hline
    \rowcolor{mygreen}\multicolumn{7}{c}{\textit{Open-weight models}}  \\ \hline
    \ssc{Llama-3.2-3B} & December 1, 2023 & \chat & 0.0 & 0.0 & 1.6 & 3.5          \\
    \ssc{Llama-3.1-70B} & December 2023 & \chat & 0.0 & 0.0 & 0.0 & 6.3          \\ 
     \ssc{Llama-4-Scout-17B-16E (109B)} & August 2024 & \chat & 0.0 & 0.0 & 5.9 & 5.9          \\ 
     \ssc{Qwen3-235B-A22B} & \multicolumn{1}{c}{-} & \reason & 0.0 & 5.4 & 4.3 & 11.4          \\ 
     \ssc{DeepSeek-R1-Distill-Qwen-1.5B} & \multicolumn{1}{c}{-} & \reason & 0.0 & 2.7 & 0.0 & 1.6          \\ 
     \ssc{DeepSeek-R1-Distill-Qwen-14B} & \multicolumn{1}{c}{-} & \reason & 0.9 & 3.6 & 0.9 &  10.6         \\ 
     \ssc{\deepseek} & \multicolumn{1}{c}{-} & \reason & \textbf{5.4} & 1.8 & \textbf{22.4} & 11.0          \\
     \ssc{gpt-oss-20b-high}$^*$ & June, 2024 & \reason & 0.9 & 4.5 & 2.7 & 7.8          \\
     \ssc{gpt-oss-120b-high}$^*$ & June, 2024 & \reason & 0.9 & \textbf{7.2} & 10.6 & \textbf{16.9}          \\
     
    \bottomrule
\end{tabular}
\end{adjustbox}
\label{table:main_results}
\vspace{-2mm}
\end{table*}
\endgroup

\vspace{-2mm}
Having established \sealqa, we now set out to evaluate how well today's \llms reason over noisy search results when navigating dynamic, real-world knowledge. Our analysis reveals limitations in their ability to reconcile conflicting parametric (\emph{internal}) and retrieved (\emph{external}) knowledge.
\vspace{-2mm}
\subsection{Experiment setup}
\vspace{-2mm}
\subsubsection{\bssc{Seal-0} and \bssc{Seal-Hard}}
\vspace{-2mm}
\paragraph{Baselines:} 

We benchmarked a wide range of open-weight and proprietary models. These include chat-oriented models such as \ssc{GPT-4o}, \ssc{GPT-4.1}, their \ssc{mini} variants~\citep{openai2024gpt4o,openai2024gpt4omini,openai2025introducing}, \ssc{Llama-3.1-70B}~\citep{grattafiori2024llama}, \ssc{Llama-3.2-3B}~\citep{meta2024llama32}, and \ssc{Llama-4-Scout-17B-16E-Instruct}~\citep{meta2025llama4}; advanced reasoning models such as \ssc{o3-mini}~\citep{openai2025openaio3mini}, \deepseek, \ssc{DeepSeek-R1-Distill-Qwen-14B/1.5B}~\citep{guo2025deepseek}, and \ssc{Qwen3-235B-A22B}~\citep{yang2025qwen3}; and agentic tool-use models such as \ssc{o3} and \ssc{o4-mini}~\citep{openai2025openaio3o4mini}.\footnote{We used the \fssc{OpenAI} and \fssc{Together.AI} \fssc{APIs} for \fssc{OpenAI} and open-weight models, respectively.} After the release of \sealqa, we additionally benchmarked \ssc{GPT-5} and \ssc{GPT-5-mini}~\citep{openai2025gpt5}, and \ssc{gpt-oss-20b} and \ssc{gpt-oss-120b}~\citep{openai2025gptoss120bgptoss20bmodel}. We mainly include \sscnoxsp{GPT-5} models as a reference for current state-of-the-art performance on \sscnoxsp{SealQA}. Data contamination may exist after the release, and we cannot prevent \sscnoxsp{GPT-5} or other agentic models from accessing our dataset links.

We fed each question as a prompt into each model, using a temperature of 0 when configurable and the default value otherwise.\footnote{\fssc{OpenAI}'s \fssc{GPT-5} and \fssc{o}-series models only support a fixed temperature of 1.0.}
For models without browsing, we applied \ssc{FreshPrompt}~\citep{vu-etal-2024-freshllms} or \ssc{self-ask}~\citep{press-etal-2023-measuring} to inject \google search results into the prompt. Advanced reasoning models were evaluated under \emph{high} reasoning effort settings when configurable, unless specified otherwise.
\vspace{-2mm}
\paragraph{Human competitors:} To estimate human performance, we asked \emph{five} graduate-level \nlp researchers (not involved in annotation) to independently answer a sample of 50 \sealhard questions. They had unlimited access to \google and could use any queries they deemed useful (\emph{open search}).\footnote{Each question had a 15-minute time limit.} \begin{table}%
\vspace{-0.5em}
\centering
\caption{On \sscnoxsp{Seal-Hard}, \sscnoxsp{LLMs} tend to underperform on cross-lingual reasoning ({$\mathcal{Q}_4$}) and false-premise detection ({$\mathcal{Q}_5$}) compared to advanced reasoning ({$\mathcal{Q}_1$}), entity/event disambiguation ({$\mathcal{Q}_2$}), and temporal tracking ({$\mathcal{Q}_3$}).
}
\begin{adjustbox}{max width=0.6\textwidth}
\begin{tabular}{c l c c c c c}
\toprule
& \textbf{Model} & {$\mathcal{Q}_1$} & {$\mathcal{Q}_2$} & {$\mathcal{Q}_3$} & {$\mathcal{Q}_4$} & {$\mathcal{Q}_5$} \\
\midrule
\multirow{5}{*}{\rotatebox[origin=c]{90}{\ssc{w/o search}}}
& \ssc{GPT-4.1}       & 14.1 & 14.2 & 25.7 & 0.0 & 0.0 \\
& \ssc{o3-mini-high}       & 10.9 & 14.9 & 14.3 & 0.0 & 0.0 \\
& \ssc{o3-high}            & –    & –    & –    & –   & –   \\
& \ssc{Llama-4‑Scout} & 4.9  & 6.8  & 5.7  & 0.0 & 0.0 \\
& \ssc{DeepSeek-R1}   & 20.7 & 23.0 & 22.9 & 7.1 & 0.0 \\
\midrule
\multirow{5}{*}{\rotatebox[origin=c]{90}{\ssc{w/ search}}}
& \ssc{GPT-4.1}       & 20.1$^\dagger$ & 17.6$^\dagger$ & 25.7$^\dagger$ & 21.4$^\dagger$ & 9.1$^\dagger$ \\
& \ssc{o3-mini-high}       & 9.8 & 10.1  & 22.9 & 7.1  & 9.1 \\
& \ssc{o3-high}            & 31.0$^\dagger$ & 31.8$^\dagger$ & 45.7$^\dagger$ & 14.3$^\dagger$  & 27.3$^\dagger$ \\
& \ssc{Llama-4‑Scout} & 4.3  & 6.8  & 8.6  & 0.0  & 0.0 \\
& \ssc{DeepSeek-R1}   & 10.3 & 10.8 & 14.3 & 0.0  & 18.2 \\
\bottomrule
\end{tabular}
\end{adjustbox}
\label{table:results_question_type}
\end{table}
Additionally, after completing the open-search task, they were given five curated URLs per question: one containing the correct answer and four containing conflicting or misleading information (\emph{oracle}).
\vspace{-2mm}
\subsubsection{\bssc{LongSeal}}
\textbf{Baselines:} We benchmarked \ssc{GPT-4o-mini}, \ssc{GPT-4.1-mini}, \ssc{Llama-4-Scout-17B-16E-Instruct}, and additionally \ssc{Llama-3.2-11B-Vision}~\citep{meta2024llama32}, with context windows of 128K, 1M, 1M, and 128K tokens, respectively.

We followed~\citet{liu-etal-2024-lost} to set up a multi-document \qa task where a model receives a question and a set of documents: one \emph{gold} document that suggests the correct answer, and $k$ hard negatives. The gold document is randomly placed among the $k$ negatives. To answer correctly, the model must identify and use the gold document from its input context. We evaluated three values of $k$: 12, 20, and 30, sampled from 50 hard negatives per question. This setup allows us to assess how performance varies with the number of negatives and the position of the gold document.\footnote{The average prompt lengths across all examples are 27.6K, 54.5K, and 70.1K tokens, with 100\%, 99.2\%, and 96.7\% of prompts fitting within the 128K context window of \fssc{GPT-4o-mini} and \fssc{LLaMA-3.2-11B}, for $k=$ 12, 20, and 30, respectively.}

\begin{table}[t]%
\centering
\caption{Questions that involve rapidly changing information, i.e., fast-changing questions, pose significant challenges for \sscnoxsp{LLMs} on \sscnoxsp{Seal-Hard}.}
\begin{adjustbox}{max width=0.6\textwidth}
\begin{tabular}{lcccccc}
\toprule
\multirow{2}{*}{\textbf{Model}} & \multicolumn{3}{c}{\ssc{w/o search}} & \multicolumn{3}{c}{\ssc{w/ search}} \\ \cline{2-7}
 & \ssc{never} & \ssc{slow} & \ssc{fast} & \ssc{never} & \ssc{slow} & \ssc{fast} \\
\midrule
\ssc{GPT-4.1}         & 21.5 & 18.0 & 1.6 & 17.7$^\dagger$ & 24.3$^\dagger$ & 17.2$^\dagger$ \\
\ssc{o3-mini-high}          & 20.3 & 12.6 & 3.1 & 12.7 & 10.8 & 10.9 \\
\ssc{o3-high}         &   –  &   –  &  –  & 39.2$^\dagger$ & 36.9$^\dagger$ & 17.2$^\dagger$ \\
\ssc{Llama-4‑Scout}                   & 10.1 &  4.5 & 4.1 &  6.3 &  4.5 & 7.8 \\
\ssc{DeepSeek-R1}                     & 32.9 & 24.3 & 6.2 & 15.2 &  9.9 & 7.8 \\
\bottomrule
\end{tabular}
\end{adjustbox}
\label{table:results_freshness}
\end{table}

\vspace{-2mm}
\subsection{Results on \bssc{Seal-0} and \bssc{Seal-Hard}}
\vspace{-2mm} 
\textbf{\bssc{Seal-0} and \bssc{Seal-Hard} present significant challenges for frontier \bssc{LLMs}:} 
Table~\ref{table:main_results} shows the accuracy of various \llms on \sealzero and \sealhard without access to a search engine (\ssc{w/o search}). %
Excluding \ssc{GPT-5} variants, all other models perform poorly without web access, with accuracies ranging from 0.0\% to 5.4\% on \sealzero and 0.0\% to 22.4\% on \sealhard.\begin{table}[t]%
\centering
\caption{\sscnoxsp{LLMs} struggle with questions that involve recent information on \sscnoxsp{Seal-Hard}.}
\begin{adjustbox}{max width=0.6\textwidth}
\begin{tabular}{lcccccc}
\toprule
\multirow{2}{*}{\textbf{Model}} & \multicolumn{3}{c}{\ssc{w/o search}} & \multicolumn{3}{c}{\ssc{w/ search}} \\ \cline{2-7}
 & \ssc{$<$ 2024} & \ssc{2024} & \ssc{2025} & \ssc{$<$ 2024} & \ssc{2024} & \ssc{2025} \\
\midrule
\ssc{GPT-4.1}          & 23.5 & 6.1 & 0.0 & 25.5$^\dagger$ & 20.4$^\dagger$ & 7.1$^\dagger$ \\
\ssc{o3-mini-high}          & 20.5 & 2.6 & 1.4 & 14.4 & 12.8 & 4.3 \\
\ssc{o3-high}         &   –  &  –  &  –  & 45.9$^\dagger$ & 15.4$^\dagger$ & 14.5$^\dagger$ \\
\ssc{Llama-4‑Scout}                   &  8.7 & 4.1 & 0.0 &  7.4 &  6.1 & 1.8 \\
\ssc{DeepSeek-R1}                    & 35.6 & 8.2 & 0.0 & 14.8 &  6.1 & 5.4 \\
\bottomrule
\end{tabular}
\end{adjustbox}
\label{table:results_effective_year}
\end{table}

While proprietary models tend to outperform open-weight ones, \deepseek stands out as a notable exception, achieving 5.4\% accuracy. Interestingly, model size does not consistently correlate with performance. For example, both \ssc{Llama-3.2-3B} and \ssc{Llama-3.1-70B} score 0.0\% on \sealzero, with the smaller model slightly outperforming the larger one on \sealhard (1.6\% vs. 0.0\%). A similar pattern holds for \ssc{DeepSeek-R1-Distill-Qwen}, which shows negligible improvement when scaled from 1.5B to 14B (0.0\% $\rightarrow$ 0.9\%) on both datasets. Large \emph{mixture-of-expert} (\ssc{MoE}) models such as \ssc{Llama-4-Scout-17B-16E} (109B total parameters) and \ssc{Qwen3-235B-A22B} also fail to generalize on \sealzero (0.0\%) and yield only modest gains on \sealhard (5.9\% and 4.3\%, respectively). Additionally, reasoning-focused models do not consistently outperform general-purpose chat models, as seen with \ssc{Qwen3-235B-A22B} and \ssc{Llama-4-Scout-17B-16E}, with \deepseek as the exception.

Tables~\ref{table:results_question_type},~\ref{table:results_freshness},~\ref{table:results_effective_year}, and~\ref{table:results_search_quality} show a breakdown of \sealhard results by question category (see Appendix~\ref{section:results_by_question_category} for full results). %
Overall, all models show limitations across question categories, especially on cross-lingual reasoning, false-premise detection, and questions that involve recent or rapidly changing information. \begin{table}[t]%
\centering
\caption{On \sscnoxsp{Seal-Hard}, performance degrades more when search results are uniformly unhelpful than when they contain conflicting answers.}
\begin{adjustbox}{max width=0.6\textwidth}
\begin{tabular}{lcccc}
\toprule
\multirow{2}{*}{\textbf{Model}} &
\multicolumn{2}{c}{\ssc{w/o search}} &
\multicolumn{2}{c}{\ssc{w/ search}} \\ \cline{2-5}
 & \ssc{unhelpful} & \ssc{conflict.} & \ssc{unhelpful} & \ssc{conflict.} \\
\midrule
\ssc{GPT-4.1}          & 14.5 & 15.3 & 18.2$^\dagger$ & 22.2$^\dagger$ \\
\ssc{o3-mini-high}          & 10.9 & 13.9 & 8.2 & 13.9 \\
\ssc{o3-high}         &   –  &   –  & 30.0$^\dagger$ & 34.7$^\dagger$ \\
\ssc{Llama-4‑Scout}                   & 3.6 &  7.6 &  4.5 &  6.9 \\
\ssc{DeepSeek-R1}                    &20.9 & 23.6 &  9.1 & 12.5 \\
\bottomrule
\end{tabular}
\end{adjustbox}
\label{table:results_search_quality}
\end{table}

\begin{figure*}
\vspace{-1.5em}
\centering
\includegraphics[width=0.7\textwidth]{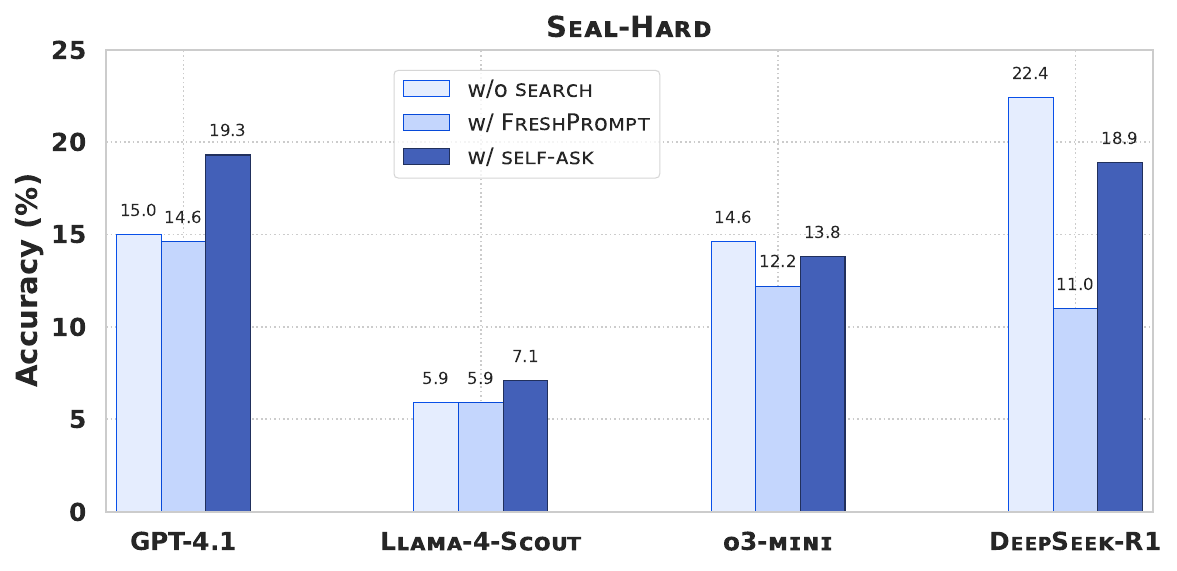}
\caption{Advanced reasoning models such as
\sscnoxsp{DeepSeek-R1-671B} and \sscnoxsp{o3-mini} are highly vulnerable to noisy search results.}
\label{figure:retrieval_based_prompting_fig}
\vspace{-0.8em}
\end{figure*}

Performance also degrades more when search results are unhelpful than when they contain conflicting answers.\footnote{Additionally, we find that open-weight models like \fssc{Llama-4-Scout} and \fssc{DeepSeek-R1} choose to ``not attempt'' questions more often than proprietary models such as \fssc{GPT-4.1}, \fssc{o4-mini}, and \fssc{o3}(see Appendix~\ref{section:results_by_answer_type}).}
\vspace{-2mm}
\paragraph{Naive search and integration can amplify noise rather than improve accuracy:} 
Table~\ref{table:main_results} (\ssc{w/o search}) and Figure~\ref{figure:retrieval_based_prompting_fig} show the effects of web search on model performance. In general, search improves accuracy across models. Agentic reasoning models such as \ssc{o3} and \ssc{o4-mini}, which can use tools within \ssc{ChatGPT} including web search, perform significantly better than others. \ssc{o3} achieves 14.4\% on \sealzero and 32.7\% on \sealhard.

\vspace{-2mm}
Our results suggest that training models to understand and execute search queries, as done in \ssc{ChatGPT}'s built-in search, is more effective than retrieval-based prompting methods like \freshprompt. While \ssc{GPT-4.1} gains a performance boost from built-in search (+5.5\%), \freshprompt slightly reduces its accuracy (15.0\% $\rightarrow$ 14.6\%). Built-in search generally improves performance on \sealhard for both \ssc{GPT-4.0} and \ssc{GPT-4.1}. With that said, \freshprompt remains useful for most open-weight models without tool-use training. For example, \ssc{Qwen3-235B-A22B} and \ssc{DeepSeek-R1-Distill-Qwen-14B} achieve gains of +7.1\% and +9.7\%, respectively, on \sealhard when using \freshprompt.

\vspace{-2mm}
However, search can sometimes be detrimental. \ssc{GPT-4.1-mini}, when equipped with built-in search, drops in accuracy from 13.8\% to 11.8\%. Since \sealqa questions are designed to elicit conflicting or noisy search results, naive retrieval and integration can harm model accuracy.

\vspace{-2mm}

\paragraph{Advanced reasoning models can be highly vulnerable to noisy search results:} As shown in Table~\ref{table:main_results} (\ssc{w/o search}) and Figure~\ref{figure:retrieval_based_prompting_fig}, \deepseek and \ssc{o3-mini} are dramatically more sensitive to input noise than other models. For example, \deepseek's performance drops from 22.4\% to 11.0\% when using \freshprompt. Our ablation (Table~\ref{table:results_freshness} and Table~\ref{table:results_effective_year}) reveals that \freshprompt improves \deepseek's performance on fast-changing (+1.6\%) and 2025-specific (+5.4\%) questions, but leads to large drops on static or older questions (-17.7\% on never-changing, and -20.8\% on pre-2024). \ssc{GPT-4.1-mini} shows a similar trend with \ssc{ChatGPT}'s built-in search, though the decline is less pronounced. In contrast, open-weight models with weaker reasoning capabilities (e.g., \ssc{Qwen3-235B-A22B} and \ssc{DeepSeek-R1-Distill-Qwen-14B}) consistently benefit from \freshprompt. 

Among retrieval-based prompting methods, \selfask, which decomposes questions into sub-questions, is generally more effective than \freshprompt, which issues direct searches and thus triggers more noise for \sealqa's adversarial questions. However, both methods harm the accuracy of \deepseek and \ssc{o3-mini}.

\vspace{-2mm}
\paragraph{Test-time scaling does not lead to reliable gains on \bssc{SealQA}:} %
Models like \ssc{GPT-5} and the \ssc{o}-series have shown strong reasoning capabilities, with consistent improvements from increased test-time compute. However, we find that this approach does not yield reliable gains on \sealqa.

\vspace{-2mm}
Figure~\ref{figure:test_time_scaling_seal0} illustrates test-time scaling effects on \sealzero questions across different reasoning effort settings: \emph{low}, \emph{medium}, and \emph{high}, where higher levels correspond to more reasoning tokens. \ssc{o3-mini}'s accuracy plateaus despite scaling, with scores of 1.8\%, 2.7\%, and 1.8\% at low, medium, and high effort levels, respectively. \ssc{o4-mini}'s accuracy peaks at low effort (6.3\%), but drops with more compute at medium (5.4\%) and high (4.5\%) settings. While \ssc{o3} achieves the highest overall accuracy among the \ssc{o}-series, scaling also fails to provide reliable gains, with accuracies of 11.7\%, 17.1\%, and 14.4\% across the three effort levels. Similar trends are observed on the latest \ssc{GPT-5} models.

\vspace{-2mm}
We conjecture that increased reasoning over noisy search results may impair performance. As test-time compute grows, longer chains of thought can amplify spurious or irrelevant information, which entangles the model in misleading evidence and ultimately reduces accuracy.
\vspace{-2mm}
\paragraph{The effect of repeated sampling:} We also examined the effect of repeated sampling~\citep{brown2024large}. Each model was sampled \emph{five} times, and an answer was counted as correct if any attempt was correct. Due to \ssc{o3}'s high \api cost, this experiment was restricted to \ssc{o3-mini} and \ssc{o4-mini}, evaluated on \sealzero at medium reasoning effort. In this setting, \begin{figure*}
\centering
\includegraphics[width=\linewidth]{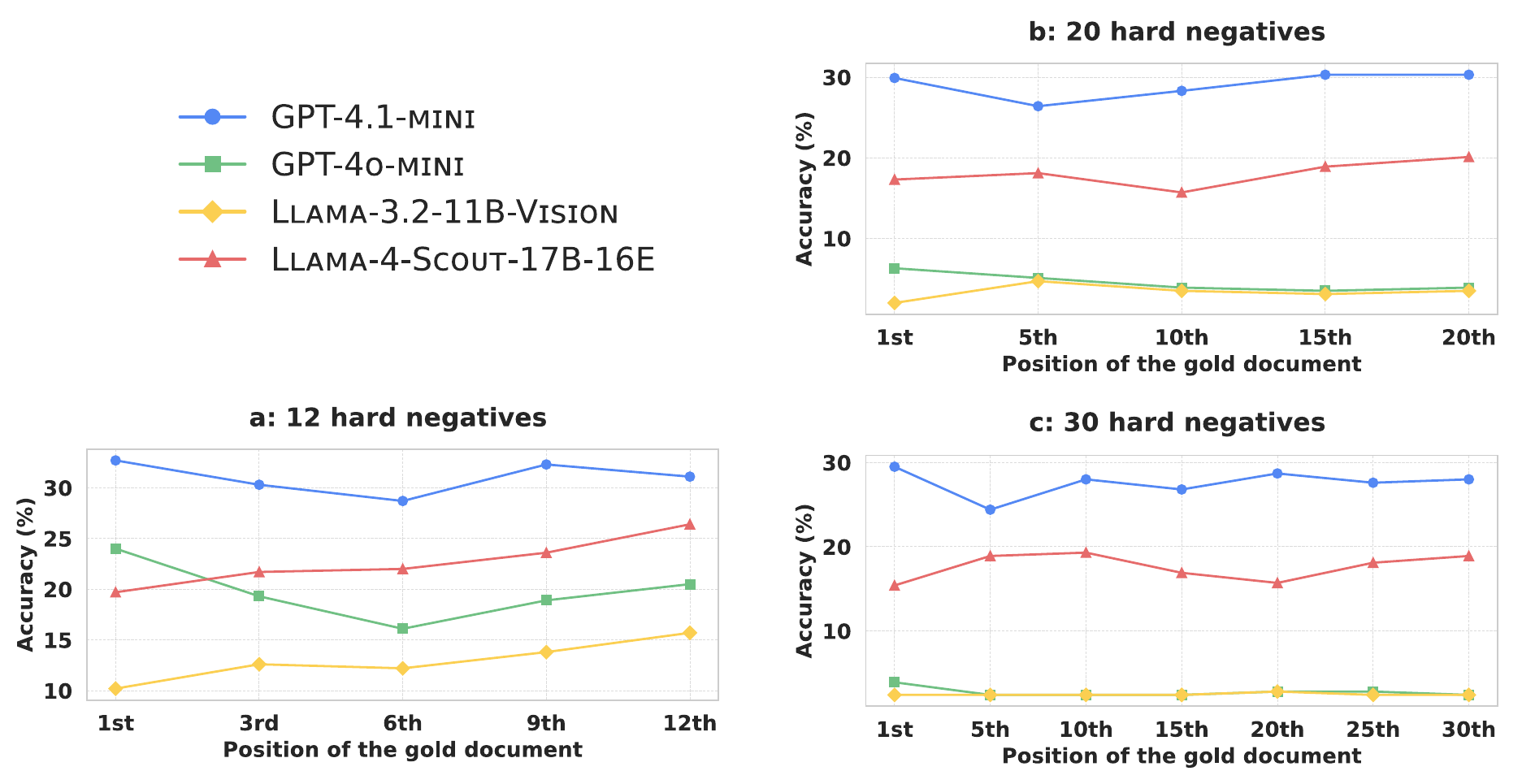}
\caption{Frontier \sscnoxsp{LLMs} fail to reliably identify relevant documents in \sscnoxsp{LongSeal} when numerous distractors are present, despite being less prone to \emph{``lost-in-the-middle''} failures~\citep{liu-etal-2024-lost}.}
\label{figure:long_context_fig}
\end{figure*}
\ssc{o3-mini} and \ssc{o4-mini} achieved 9\% and 16.2\% accuracy, respectively. These results again show that \sealzero is extremely challenging, even for agentic reasoning models with full tool access.
\vspace{-2mm}
\paragraph{\bssc{SealQA} requires careful search and robust reasoning:} 
Table~\ref{table:human_performance} in Appendix~\ref{section:appendix_human_performance} shows that frontier LLMs lag behind humans on \sealqa: the best model, \ssc{o3-high}, reached 28.0\% accuracy, compared with human averages of 38.8\% in open search and 50.4\% in oracle, and top human scores of 64.0\% and 72.0\%, respectively. Humans answered within five minutes in 52.8\% of cases but were correct only 53.0\% of the time, which highlights the dual challenge of retrieving relevant information and reasoning through conflicting sources in \sealqa.
\vspace{-2mm}
\subsection{Results on \bssc{LongSeal}}
We now switch gears to discuss our evaluation results on \longseal (Figure~\ref{figure:long_context_fig}).

\vspace{-2mm}

\paragraph{Frontier \bssc{LLMs} struggle on \bssc{LongSeal} with increased distractors:} 

All models exhibit a clear drop in accuracy as the number of hard negatives increases. For example, when the gold document appears immediately after the question (1$^{st}$ position), \ssc{GPT-4.1-mini}'s accuracy decreases from 32.7\% at $k=12$ (12 hard negatives, Figure~\ref{figure:long_context_fig}a) to 29.9\% at $k=20$ and 29.5\% at $k=30$ (Figure~\ref{figure:long_context_fig}b and c, respectively). The degradation is more pronounced in smaller or less capable models: \ssc{GPT-4o-mini} falls from 24.0\% to 6.3\% and then 3.9\%, while \ssc{LLaMA-3.2-11B} drops from 10.2\% to 2.0\% and 2.4\%.

These results indicate that simply increasing context size does not guarantee effective context use. When many hard negatives are present, models often struggle to identify and prioritize the gold document. The primary failure mode appears to be the inability to reliably filter relevant from irrelevant content at scale. High distractor density impairs relevance estimation, even when all input documents fit within the context window. This suggests a need for architectural advances or training strategies that enhance implicit retrieval and salience detection to improve performance in large-context, multi-document \qa settings.

\vspace{-2mm}
To understand why models achieve low performance, we conducted an experiment to disentangle reasoning from retrieval. In this setup, models received only gold documents, with hard negatives removed. Table~\ref{tab:only_golds} shows that performance remains low even under these ideal conditions (\ssc{w / evidence}). \ssc{GPT-4.1} and \ssc{o3-mini} achieved only 48.0\% and 56.7\%, respectively, and no open-source models exceeded 50.0\%. These results suggest that \sealqa's difficulty arises from both reasoning and retrieval challenges.

\vspace{-2mm}
\paragraph{Absence of classic positional bias in~\citet{liu-etal-2024-lost}:}
Unlike earlier work that reports a strong \emph{``lost in the middle''} effect, our results show no clear U-shaped positional trend. \ssc{GPT-4.1-mini} maintains stable accuracy across positions, with only minor fluctuations from start to end; even at $k=30$, its performance varies little between early, middle, and late placements. \ssc{LLaMA-4-Scout} shows a slight improvement toward later positions, but no consistent dip in the middle.

This absence of positional bias suggests that newer models may have mitigated some of the structural weaknesses previously associated with position encoding. However, the broader challenge remains: regardless of position, models often fail to recognize the gold document when distractors are numerous. The issue has shifted from sensitivity to position to a more general difficulty in modeling relevance within large, noisy contexts.
\begin{table}%
\vspace{-2em}
    \centering
    \caption{Frontier models fail to extract correct answers even when no distractors are provided.%
    }
    \resizebox{0.5\linewidth}{!}{%
        \begin{tabular}{l ccc}
        \toprule
            \textbf{Models} & \ssc{w/ evidence} & \ssc{w/o search} & \ssc{w/ search} \\
        \midrule
            \ssc{GPT-4.1} & 48.0 & 15.0 & 20.5 \\
            \ssc{o3-mini} & 56.7 & 14.6 & 12.2 \\
            \ssc{Llama-4-Scout} & 33.5 & 5.9 & 5.9 \\
            \ssc{DeepSeek-R1} & 49.2 & 19.3 & 15.4 \\
        \bottomrule
        \end{tabular}
    }
    \label{tab:only_golds}
    \vspace{-1em}
\end{table}

\vspace{-2.2mm}
\section{Related work}
\vspace{-2.8mm}
\paragraph{Reasoning under knowledge conflict:}\ Prior work shows that \llms can be vulnerable to misinformation~\citep{pan-etal-2023-attacking}, irrelevant context~\citep{pmlr-v202-shi23a}, and conflicting sources~\citep{NEURIPS2023_7adce80e}. Retrieval quality strongly influences model output; however, contradictions between sources often have only a minimal effect on model confidence~\citep{chen-etal-2022-rich}. ~\citet{wan-etal-2024-evidence} find that models prioritize surface-level relevance over credibility indicators such as scientific references or neutral tone. While \llms can detect conflict~\citep{jiayang-etal-2024-econ}, they struggle to resolve it~\citep{wang2024resolving,xu2024debateqa}. Models also exhibit \emph{confirmation bias} by favoring evidence that aligns with their parametric memory~\citep{chen-etal-2022-rich}, often resolving contradictions in favor of internal knowledge~\citep{jin-etal-2024-tug,jiayang-etal-2024-econ}. Still, ~\citet{xie2024adaptive} show that models remain highly receptive to contradictory external evidence when it is coherent and convincing. Additional biases include favoring frequent evidence and relying on memory for common knowledge but external sources for long-tail knowledge~\citep{jin-etal-2024-tug}. %
See~\citet{xu-etal-2024-knowledge-conflicts} for a comprehensive survey.

\vspace{-2mm}
Building on these insights, recent work has introduced benchmarks targeting specific types of retrieval conflicts. Some focus on specific challenges, such as entity ambiguity~\citep[\ssc{AmbigDocs;}][]{lee2024ambigdocs}, credible yet conflicting sources~\citep[\ssc{WikiContradict};][]{NEURIPS2024_c6381975}, debatable questions %
~\citep[\ssc{DebateQA};][]{xu2024debateqa}, and~\citet{shaier-etal-2024-adaptive} for citation-aware \qa under ambiguity. Other assess model behavior under noisy contexts, such as faithfulness under unanswerable, inconsistent, and counterfactual contexts~\citep[\ssc{FaithEval};][]{ming2025faitheval}, or reasoning over conflicting contexts~\citep[\ssc{QACC};][]{liu-etal-2025-open}, as well as analyzing what shapes predictions, such as textual features~\citep[\ssc{ConflictingQA};][]{wan-etal-2024-evidence} and conflict sources~\citep[\ssc{ConflictBank};][]{NEURIPS2024_baf4b960}. Most recently,~\citet{wang2025retrieval} augment \ssc{AmbigDocs} examples with simulated ambiguity, misinformation, and noise to create \ssc{RAMDocs}. Our work complements this growing body by introducing a unified benchmark that brings together real-world challenges, such as ambiguity, misinformation, temporal drift, and noisy retrieval, through expert-curated, naturally occurring questions, without relying on synthetic augmentation. %

\vspace{-2.5mm}
\paragraph{Measuring factuality and reasoning in \bssc{LLMs}:}\ 
\sealqa aligns with a growing body of work on time-sensitive \qa benchmarks~\citep[][\textit{inter alia}]{NEURIPS_DATASETS_AND_BENCHMARKS2021_1f0e3dad,zhang-choi-2021-situatedqa,pmlr-v162-liska22a,NEURIPS2023_9941624e,vu-etal-2024-freshllms}. \sealqa also fits among recent \emph{challenging} benchmarks that evaluate \llms  across factuality, reasoning, and retrieval. Benchmarks like \ssc{MMLU}~\citep{hendrycks2021measuring}, \ssc{MATH}~\citep{NEURIPS_DATASETS_AND_BENCHMARKS2021_be83ab3e}, \ssc{GPQA}~\citep{rein2024gpqa}, and \ssc{Humanity's Last Exam}~\citep{phan2025humanity} focus on academic or expert-level reasoning. Others evaluate open-domain retrieval~\citep[\ssc{FreshStack};][]{thakur2025freshstack}, multi-hop, multi-document reasoning~\citep[\ssc{FRAMES};][]{krishna-etal-2025-fact}, and real-world software engineering tasks~\citep[\ssc{SWE-bench};][]{jimenez2024swe}. Targeted evaluations such as \ssc{SimpleQA}~\citep{wei2024measuring} and \ssc{BrowseComp}~\citep{wei2025browsecomp} measure factual recall and web browsing competence. These datasets push different axes of model performance, and \sealqa complements them by providing a unified benchmark spanning all three dimensions: factuality, reasoning, and retrieval, through naturally occurring, adversarially curated questions that reflect real-world \qa complexity.

\vspace{-2.2mm}
\section{Conclusion}
\vspace{-2.8mm}
We introduce \sealqa, a benchmark for evaluating Search-Augmented Language Models on challenging factual questions where web search results may be conflicting, noisy, or irrelevant. \sealqa includes three flavors: \sealzero, which includes questions that challenge today's frontier models; \sealhard, a wider collection of difficult queries; and \longseal, which is designed to test long-context reasoning in ``needle-in-a-haystack'' settings. Our evaluations show that frontier \llms, including agentic models with search tools, underperform on \sealqa and are vulnerable to noisy search results, with increased test-time compute often not leading to reliable performance gains. \longseal in particular highlights the difficulty models face in identifying relevant information amid distractors, though they exhibit reduced susceptibility to the ``lost-in-the-middle'' issue. We hope that \sealqa will spur more fundamental research into tackling real-world challenges in retrieval-augmented reasoning.

\newpage
\section*{Code of ethics and ethics statement}
We ensure that all sources used in \sealqa are publicly available and used exclusively for academic research in full compliance with the copyright terms of the original sources. We carefully verify that none of the data include harmful content such as racial discrimination, violence, or private information. The dataset is freely available to researchers for academic purposes. All data and experiments presented in our work follow scientific standards that guarantee the authenticity and accuracy of the results.

\section*{Reproducibility}
The datasets and annotation process are detailed in Section \ref{sec:data}, and the experimental settings are presented in Section \ref{sec:exp}.

\section*{Acknowledgments}
We acknowledge Advanced Research Computing at Virginia Tech for providing computational resources and support\footnote{\url{https://arc.vt.edu/}}.
We thank Quyet Do, Amartya Dutta, and Rishab Balasubramanian for their contributions to creating and reviewing \sealqa questions. We are also grateful to Yeana Bond, Pin-Jie Lin, Rishab Balasubramanian, Rituraj Sharma, Kavana Venkatesh, Lewis Bass, and Jing Chen for their involvement in the human performance evaluation. We thank the \ssc{VT LLMs} group for valuable discussions and feedback. Finally, we thank Arie Cattan, Mohit Iyyer, and Marzena Karpinska for their helpful suggestions. This research was supported by a research award from Google DeepMind and Google Research.
\clearpage

\bibliography{iclr2026/iclr2026_conference}
\bibliographystyle{iclr2026_conference}

\clearpage
\appendix
\section{Additional test-time scaling results on \textsc{SealQA}}
\begin{figure}[h!]
\centering
\includegraphics[width=\textwidth]{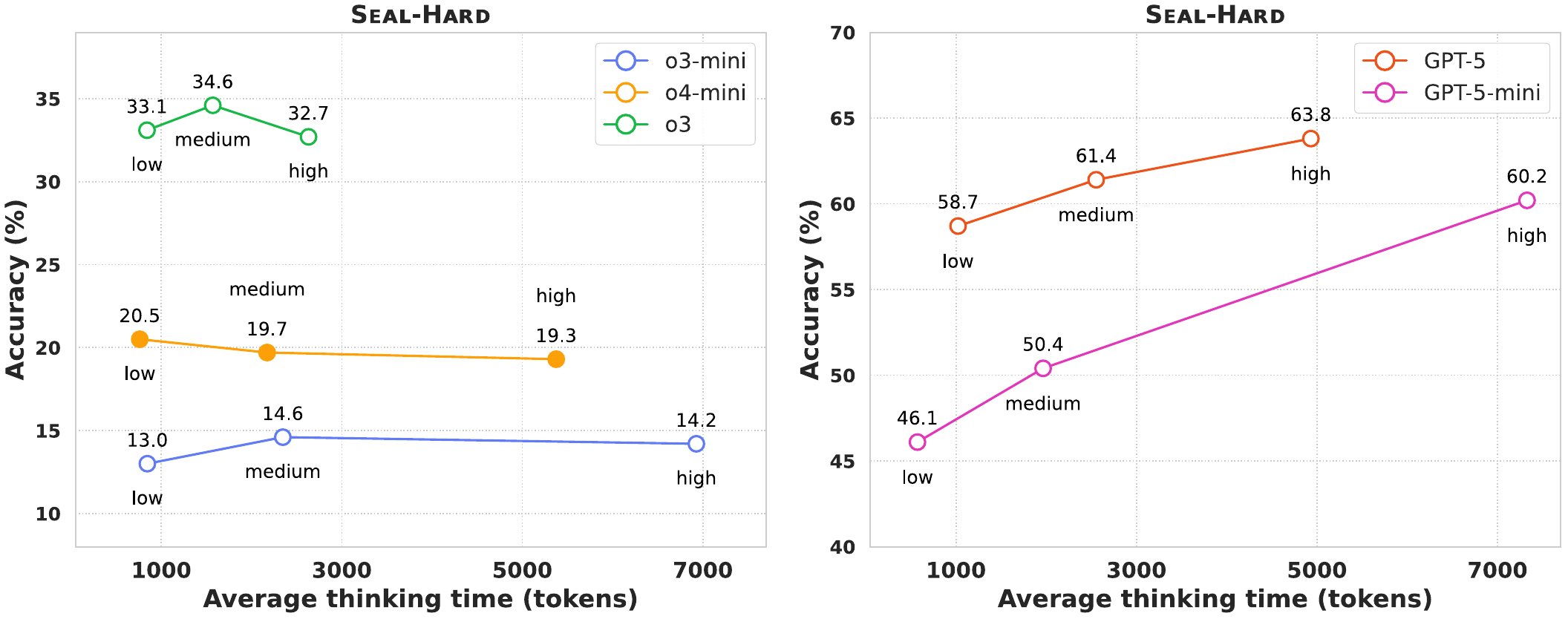}
\caption{Frontier model performance before (left) and after (right) the release of \sealhard. Despite potential data contamination or direct access by agentic models, \sealhard continues to pose a significant challenge for current frontier LLMs.}
\label{figure:test_time_scaling_sealhard}
\vspace{-2mm}
\end{figure}

\label{section:appendix_test_time_scaling}
\newpage
\section{Sample \textsc{SealQA} questions}
\label{section:appendix_sample_questions}
\begin{figure*}[h]
\centering
\includegraphics[width=\textwidth]{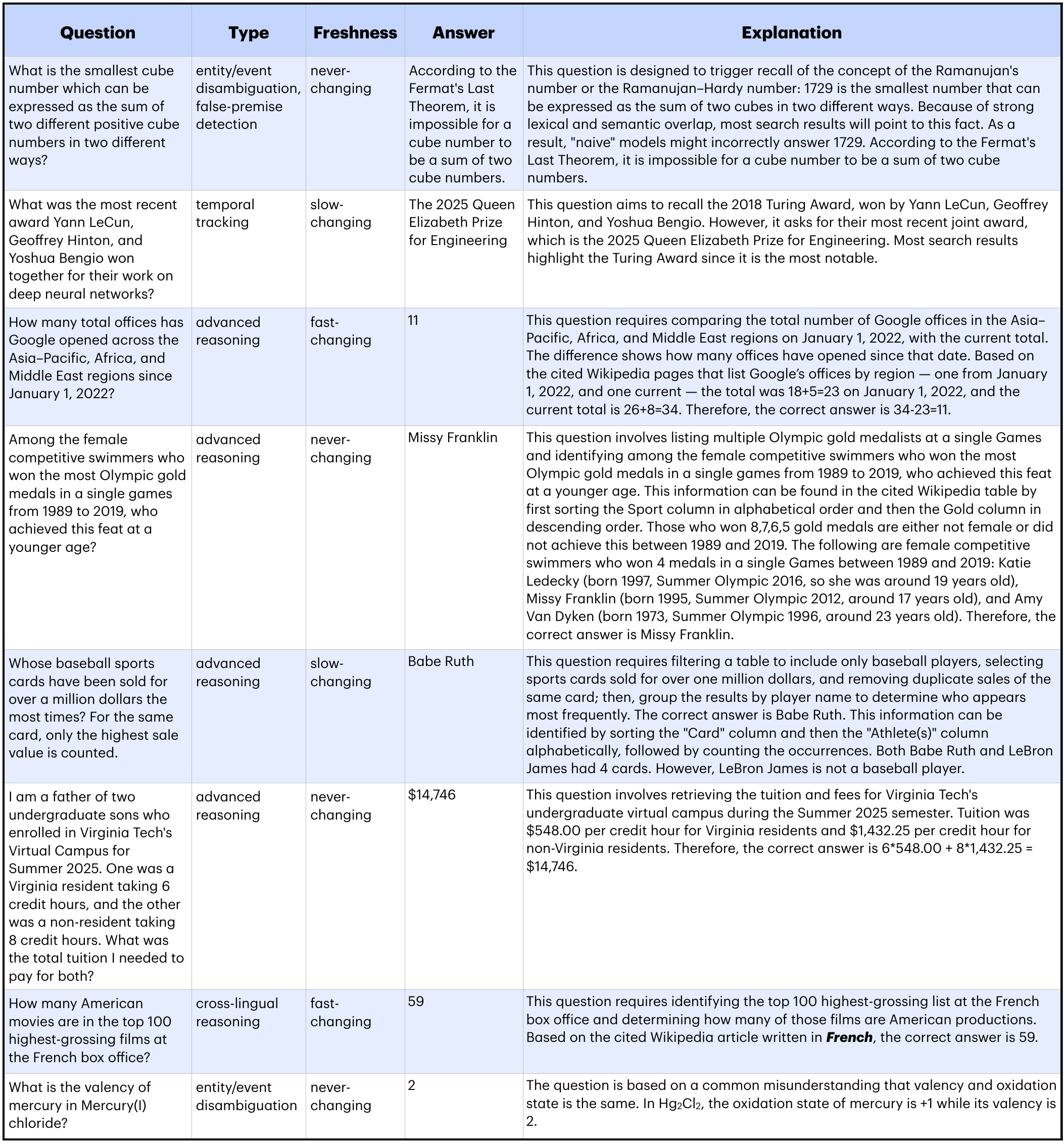}
\caption{\sealqa questions test a broad range of reasoning skills that are often overlooked in existing benchmarks.}
\label{figure:sealqa_sample_examples_fig}
\vspace{-2mm}
\end{figure*}

\newpage
\section{Template for \textsc{GPT-4o mini} auto-rater}
\label{section:grader_template} 
\begin{lstlisting}[language={},frame=single,breaklines=true]
GRADER_TEMPLATE = """
Your job is to look at a question, a gold target, and a predicted answer, and then assign a grade of either ["CORRECT", "INCORRECT", "NOT_ATTEMPTED"].
First, I will give examples of each grade, and then you will grade a new example.


The following are examples of CORRECT predicted answers.
```
Question: What are the names of Barack Obama's children?
Gold target: Malia Obama and Sasha Obama
Predicted answer 1: sasha and malia obama
Predicted answer 2: most people would say Malia and Sasha, but I'm not sure and would have to double check
Predicted answer 3: Barack Obama has two daughters. Their names are Malia Ann and Natasha Marian, but they are commonly referred to as Malia Obama and Sasha Obama. Malia was born on July 4, 1998, and Sasha was born on June 10, 2001.
```
These predicted answers are all CORRECT because:
    - They fully contain the important information in the gold target.
    - They do not contain any information that contradicts the gold target.
    - Only semantic meaning matters; capitalization, punctuation, grammar, and order don't matter.
    - Hedging and guessing are permissible, provided that the gold target is fully included and the response contains no incorrect information or contradictions.


The following are examples of INCORRECT predicted answers.
```
Question: What are the names of Barack Obama's children?
Gold target: Malia and Sasha
Predicted answer 1: Malia.
Predicted answer 2: Malia, Sasha, and Susan.
Predicted answer 3: Barack Obama does not have any children.
Predicted answer 4: I think it's either Malia and Sasha. Or it could be Malia and Jackie. Or it could be Joey and Malia.
Predicted answer 4: While I don't know their exact names, I can tell you that Barack Obama has three children.
Predicted answer 5: It's possible you may mean Betsy and Olivia. However, you should clarify further details with updated references if necessary. Is that the correct answer?
Predicted answer 6: It may be the case that Obama's child is named James. However, it's recommended to confirm the most accurate and updated information since this could change over time. This model may not always reflect the most current information.
```
These predicted answers are all INCORRECT because:
    - A factual statement in the answer contradicts the gold target. Incorrect statements that have some hedging (e.g., "it is possible that", "although i'm not sure, i think") are also considered incorrect.


The following are examples of NOT_ATTEMPTED predicted answers.
```
Question: What are the names of Barack Obama's children?
Gold target: Malia and Sasha
Predicted answer 1: I don't know.
Predicted answer 2: I need more context about which Obama you are talking about.
Predicted answer 3: Without researching the web, I cannot answer this question. However, I can tell you that Barack Obama has two children.
Predicted answer 4: Barack Obama has two children. I know that one of them is Malia, but I'm not sure about the other one.
```
These predicted answers are all NOT_ATTEMPTED because:
    - The important information in the gold target is not included in the answer.
    - No statements in the answer contradict the gold target.


Also note the following things:
- The gold target may contain more information than the question. In such cases, the predicted answer only needs to contain the information that is in the question.
    - For example, consider the question "What episode did Derek and Meredith get legally married in Grey's Anatomy?" with gold target "Season 7, Episode 20: White Wedding". Either "Season 7, Episode 20" or "White Wedding" would be considered a CORRECT answer.
- Do not punish predicted answers if they omit information that would be clearly inferred from the question.
    - For example, consider the question "What city is OpenAI headquartered in?" and the gold target "San Francisco, California". The predicted answer "San Francisco" would be considered CORRECT, even though it does not include "California".
    - Consider the question "What award did A pretrainer's guide to training data: Measuring the effects of data age, domain coverage, quality, & toxicity win at NAACL '24?", the gold target is "Outstanding Paper Award". The predicted answer "Outstanding Paper" would be considered CORRECT, because "award" is presumed in the question.
- Do not give credit for an answer if it contains any internal inconsistency.
    - For example, consider the question: "How many NBA players have scored 60 or more points in a regular season game since 2024?" with the gold answer "8". A response is INCORRECT if it states "8 players" but lists 7 or 9, or if it initially says "8 players" but later contradicts this by concluding 7 or 9.


Here is a new example. Simply reply with either CORRECT, INCORRECT, NOT ATTEMPTED. Don't apologize or correct yourself if there was a mistake; we are just trying to grade the answer.
```
Question: {question}
Gold target: {target}
Predicted answer: {predicted_answer}
```

Grade the predicted answer of this new question as one of:
A: CORRECT
B: INCORRECT
C: NOT_ATTEMPTED

Just return the letters "A", "B", or "C", with no text around it.
""".strip()
\end{lstlisting}

\newpage
\section{\textsc{Seal-Hard} results by question category}
\label{section:results_by_question_category}
Tables~\ref{table:results_question_type_without_search_full},~\ref{table:results_question_type_with_search_full},~\ref{table:results_freshness_full},~\ref{table:results_effective_year_full}, and~\ref{table:results_search_quality_full} show a breakdown of \sealhard results by question category. Overall, models perform poorly across question categories, especially on cross-lingual reasoning, false-premise detection, and questions that involve recent or rapidly changing information. Performance also degrades more when search results are uniformly unhelpful than when they contain conflicting answers.

\begingroup
\renewcommand{\arraystretch}{1.3}
\begin{table*}[h]
\centering
\caption*{$^\dagger$ indicates results using \ssc{ChatGPT}'s built-in search; all other search-based results use \ssc{FreshPrompt}~\citep{vu-etal-2024-freshllms}.}
\caption{On \sealhard, \llms tend to underperform on cross-lingual reasoning ({$\mathcal{Q}_4$}) and false-premise detection ({$\mathcal{Q}_5$}) compared to advanced reasoning ({$\mathcal{Q}_1$}), entity/event disambiguation ({$\mathcal{Q}_2$}), and temporal tracking ({$\mathcal{Q}_3$}).
}
\begin{adjustbox}{max width=0.7\textwidth}
\begin{tabular}{lccccc}
\toprule
\multirow{2}{*}{\textbf{Model}} &
\multicolumn{5}{c}{\ssc{w/o search}} \\ \cline{2-6}
 & $\mathcal{Q}_1$ & $\mathcal{Q}_2$ & $\mathcal{Q}_3$ & $\mathcal{Q}_4$ & $\mathcal{Q}_5$ \\ \hline
\rowcolor{myyellow}\multicolumn{6}{c}{\textit{Closed‑source models}} \\ \hline
\ssc{GPT-4o-mini}      &  6.5 & 7.4 & 22.9 &  7.1 & 0.0 \\
\ssc{GPT-4.1-mini}     & 10.9 & 15.5 & 22.9 & 14.3 & 9.1 \\
\ssc{GPT-4o}           & 9.8 & 13.5 & 11.4 &  0.0 & 0.0 \\
\ssc{GPT-4.1}          & 14.1 & 14.2 & 25.7 & 0.0 & 0.0 \\
\ssc{o3-mini-high}          & 10.9 & 14.9 & 14.3 & 0.0 & 0.0 \\
\ssc{o4-mini-high}    &   –  &  –  &   –  &   –  &  –  \\
\ssc{o3-high}         &   –  &  –  &   –  &   –  &  –  \\
\ssc{GPT-5-mini-high}         &  15.2  &  18.9  &   20.0  &   0.0  &  9.1  \\
\ssc{GPT-5-high}         &   34.2  &  41.9  &   34.3  &   21.4  &  36.4  \\
\hline
\rowcolor{mygreen}\multicolumn{6}{c}{\textit{Open‑weight models}} \\ \hline
\ssc{Llama-3.2‑3B}                    &  0.0 & 1.4 &  0.0 & 0.0 & 0.0 \\
\ssc{Llama-3.1‑70B}                   &  3.3 & 4.7 &  5.7 & 0.0 & 0.0 \\
\ssc{Llama-4‑Scout-17B-16E (109B)}                   &  4.9 & 6.8 &  5.7 & 0.0 & 0.0 \\
\ssc{Qwen3-235B-A22B}                           &  2.2 & 4.1 &  5.7 & 0.0 & 0.0 \\
\ssc{DeepSeek‑R1‑Distill-Qwen‑1.5B}  &  1.1 & 2.0 &  0.0 & 0.0 & 0.0 \\
\ssc{DeepSeek‑R1‑Distill-Qwen‑14B}   &  6.5 & 8.1 & 17.1 & 0.0 & 0.0 \\
\deepseek                     & 20.7 & 23.0 & 22.9 & 7.1 & 0.0 \\
\ssc{\newdeepseek}                     & 18.5 & 19.6 & 20.0 & 7.1 & 9.1 \\
\ssc{gpt-oss-20b-high}                     & 2.2  & 3.4  & 0.0  & 0.0 & 9.1 \\
\ssc{gpt-oss-120b-high}                    & 8.7 & 13.5  & 5.7  & 0.0 & 9.1 \\
\bottomrule
\end{tabular}
\end{adjustbox}
\label{table:results_question_type_without_search_full}
\vspace{-2mm}
\end{table*}
\endgroup

\begingroup
\renewcommand{\arraystretch}{1.3}
\begin{table*}[h]
\centering
\caption*{$^\dagger$ indicates results using \ssc{ChatGPT}'s built-in search; all other search-based results use \ssc{FreshPrompt}~\citep{vu-etal-2024-freshllms}.}
\caption{On \sealhard, \llms tend to underperform on cross-lingual reasoning ({$\mathcal{Q}_4$}) and false-premise detection ({$\mathcal{Q}_5$}) compared to advanced reasoning ({$\mathcal{Q}_1$}), entity/event disambiguation ({$\mathcal{Q}_2$}), and temporal tracking ({$\mathcal{Q}_3$}).
}
\begin{adjustbox}{max width=0.65\textwidth}
\begin{tabular}{lccccc}
\toprule
\multirow{2}{*}{\textbf{Model}} &
\multicolumn{5}{c}{\ssc{w/ search}} \\ \cline{2-6}
 & $\mathcal{Q}_1$ & $\mathcal{Q}_2$ & $\mathcal{Q}_3$ & $\mathcal{Q}_4$ & $\mathcal{Q}_5$ \\ \hline

\rowcolor{myyellow}\multicolumn{6}{c}{\textit{Closed‑source models}} \\ \hline
\ssc{GPT-4o-mini}      & 11.4$^\dagger$ &  10.8$^\dagger$ & 17.1$^\dagger$ & 14.3$^\dagger$ & 9.1$^\dagger$ \\
\ssc{GPT-4.1-mini}     & 8.2$^\dagger$ &  11.5$^\dagger$ & 14.3$^\dagger$ &  0.0$^\dagger$ & 0.0$^\dagger$ \\
\ssc{GPT-4o}           & 11.4$^\dagger$ &  15.5$^\dagger$ & 17.1$^\dagger$ & 7.1$^\dagger$ & 0.0$^\dagger$ \\
\ssc{GPT-4.1}          & 20.1$^\dagger$ &  17.6$^\dagger$ & 25.7$^\dagger$ & 21.4$^\dagger$ & 9.1$^\dagger$ \\
\ssc{o3-mini-high}          &  9.8 & 10.1 & 22.9 & 7.1 & 9.1 \\
\ssc{o4-mini-high}   & 20.1$^\dagger$ & 18.2$^\dagger$ & 22.6$^\dagger$ & 0.0$^\dagger$ & 9.1$^\dagger$ \\
\ssc{o3-high}         & 31.0$^\dagger$ & 31.8$^\dagger$ & 45.7$^\dagger$ & 14.3$^\dagger$ & 27.3$^\dagger$ \\
\ssc{GPT-5-mini-high}         &   61.4$^\dagger$  &  57.4$^\dagger$  &  57.1$^\dagger$  &  57.14$^\dagger$  &  45.5$^\dagger$  \\
\ssc{GPT-5-high}         &   64.7$^\dagger$  & 60.8$^\dagger$  &  57.1$^\dagger$  &  57.1$^\dagger$  &  54.5$^\dagger$  \\ \hline

\rowcolor{mygreen}\multicolumn{6}{c}{\textit{Open‑weight models}} \\ \hline
\ssc{Llama-3.2‑3B}                    &  2.7 &  2.7 &  8.6 &  0.0 & 0.0 \\
\ssc{Llama-3.1‑70B}                   &  4.3 &  4.7 & 14.3 &  7.1 & 9.1 \\
\ssc{Llama-4‑Scout-17B-16E (109B)}                   &  4.3 &  6.8 &  8.6 &  0.0 & 0.0 \\
\ssc{Qwen3-235B-A22B}                           & 9.2 & 10.8 & 14.3 & 0.0 & 18.2 \\
\ssc{DeepSeek‑R1‑Distill-Qwen‑1.5B}  &  1.1 &  2.7 &  0.0 & 0.0 & 0.0 \\
\ssc{DeepSeek‑R1‑Distill-Qwen‑14B}   &  8.2 & 9.5 & 25.7 &  0.0 & 18.2 \\
\deepseek                     & 10.3 & 10.8 & 14.3 & 0.0 & 18.2 \\
\ssc{\newdeepseek}                     & 15.2 & 12.8 & 17.1 & 7.1 & 18.2 \\
\ssc{gpt-oss-20b-high}                     & 5.4 & 9.5  & 8.6 & 0.0 & 18.2 \\
\ssc{gpt-oss-120b-high}                     & 13.0 & 22.3 & 11.4 & 0.0 & 36.4 \\

\bottomrule
\end{tabular}
\end{adjustbox}
\label{table:results_question_type_with_search_full}
\end{table*}
\endgroup

\begingroup
\renewcommand{\arraystretch}{1.3}
\begin{table*}[h]
\centering
\caption{Questions that involve rapidly changing information, i.e., fast-changing questions, pose significant challenges for \llms on \sealhard.}
\begin{adjustbox}{max width=0.65\textwidth}
\begin{tabular}{lcccccc}
\toprule
\multirow{2}{*}{\textbf{Model}} & \multicolumn{3}{c}{\ssc{w/o search}} & \multicolumn{3}{c}{\ssc{w/ search}} \\ \cline{2-7}
 & \ssc{never} & \ssc{slow} & \ssc{fast} & \ssc{never} & \ssc{slow} & \ssc{fast} \\ \hline

\rowcolor{myyellow}\multicolumn{7}{c}{\textit{Closed‑source models}} \\ \hline
\ssc{GPT-4o-mini}      & 15.2 &  9.0 & 1.6 & 16.5$^\dagger$ & 10.8$^\dagger$ & 14.1$^\dagger$ \\
\ssc{GPT-4.1-mini}     & 20.3 & 15.3 & 3.1 & 12.7$^\dagger$ & 10.8$^\dagger$ & 12.5$^\dagger$ \\
\ssc{GPT-4o}           & 16.5 & 12.6 & 4.7 & 15.2$^\dagger$ & 15.3$^\dagger$ & 14.1$^\dagger$ \\
\ssc{GPT-4.1}         & 21.5 & 18.0 & 1.6 & 17.7$^\dagger$ & 24.3$^\dagger$ & 17.2$^\dagger$ \\
\ssc{o3-mini-high}          & 20.3 & 12.6 & 3.1 & 12.7 & 10.8 & 10.9 \\
\ssc{o4-mini-high}   &   –  &   –  &  –  & 24.1$^\dagger$ & 19.8$^\dagger$ & 12.5$^\dagger$ \\
\ssc{o3-high}         &   –  &   –  &  –  & 39.2$^\dagger$ & 36.9$^\dagger$ & 17.2$^\dagger$ \\ 
\ssc{GPT-5-mini-high}         &   27.8 &  16.2  &  4.7  & 55.7$^\dagger$ & 63.1$^\dagger$ & 60.9$^\dagger$ \\ 
\ssc{GPT-5-high}         &   48.1 &   42.3  &  17.2  & 64.6$^\dagger$ & 69.4$^\dagger$ & 53.1$^\dagger$ \\ \hline

\rowcolor{mygreen}\multicolumn{7}{c}{\textit{Open‑weight models}} \\ \hline
\ssc{Llama-3.2‑3B}                    &  1.3 &  0.9 & 0.0 &  3.8 &  4.5 & 1.6 \\
\ssc{Llama-3.1‑70B}                   &  7.6 &  2.7 & 3.1 &  6.3 &  8.1 & 3.1 \\
\ssc{Llama-4‑Scout-17B-16E (109B)}                   & 10.1 &  4.5 & 4.1 &  6.3 &  4.5 & 7.8 \\
\ssc{Qwen3-235B-A22B}                           &  7.6 &  3.6 & 1.6 & 12.7 &  8.1 &15.6 \\
\ssc{DeepSeek‑R1‑Distill-Qwen‑1.5B}  &  0.0 &  1.8 & 1.6 &  1.3 &  2.7 & 0.0 \\
\ssc{DeepSeek‑R1‑Distill-Qwen‑14B}   &  7.6 &  9.0 & 4.7 & 10.1 &  9.0 &14.1 \\
\deepseek                     & 32.9 & 24.3 & 6.2 & 15.2 &  9.9 & 7.8 \\
\ssc{\newdeepseek}                     & 31.6 & 18.0 & 6.3 & 19.0 & 14.4 & 12.5 \\
\ssc{gpt-oss-20b-high}                      & 5.1 & 2.7 & 0.0 & 11.4 & 6.3 & 6.3 \\
\ssc{gpt-oss-120b-high}                      & 19.0 & 8.1 & 4.7 & 29.1 & 13.5 & 7.8 \\
\bottomrule
\end{tabular}
\end{adjustbox}
\label{table:results_freshness_full}
\end{table*}
\endgroup

\begingroup
\renewcommand{\arraystretch}{1.3}
\begin{table*}[ht!]
\centering
\caption*{$^\dagger$ indicates results using \ssc{ChatGPT}'s built-in search; all other search-based results use \ssc{FreshPrompt}~\citep{vu-etal-2024-freshllms}.}
\caption{\llms struggle with questions that involve recent information on \sealhard.}
\begin{adjustbox}{max width=0.8\textwidth}
\begin{tabular}{lcccccc}
\toprule
\multirow{2}{*}{\textbf{Model}} & \multicolumn{3}{c}{\ssc{w/o search}} & \multicolumn{3}{c}{\ssc{w/ search}} \\ \cline{2-7}
 & \ssc{before 2024} & \ssc{2024} & \ssc{2025} & \ssc{before 2024} & \ssc{2024} & \ssc{2025} \\ \hline

\rowcolor{myyellow}\multicolumn{7}{c}{\textit{Closed‑source models}} \\ \hline
\ssc{GPT-4o-mini}      & 13.4 & 6.1 & 0.0 & 16.1$^\dagger$ & 16.3$^\dagger$ & 3.6$^\dagger$ \\
\ssc{GPT-4.1-mini}     & 20.1 & 8.2 & 1.8 & 10.7$^\dagger$ & 20.4$^\dagger$ & 7.1$^\dagger$ \\
\ssc{GPT-4o}           & 16.8 & 8.2 & 1.8 & 15.4$^\dagger$ & 18.4$^\dagger$ & 10.7$^\dagger$ \\
\ssc{GPT-4.1}          & 23.5 & 6.1 & 0.0 & 25.5$^\dagger$ & 20.4$^\dagger$ & 7.1$^\dagger$ \\
\ssc{o3-mini-high}          & 20.5 & 2.6 & 1.4 & 14.4 & 12.8 & 4.3 \\
\ssc{o4-mini-high}   &   –  &  –  &  –  & 26.7$^\dagger$ & 7.7$^\dagger$ & 10.1$^\dagger$ \\
\ssc{o3-high}         &   –  &  –  &  –  & 45.9$^\dagger$ & 15.4$^\dagger$ &14.5$^\dagger$ \\
\ssc{GPT-5-mini-high}         &   26.7  &  2.6  &  4.3  & 58.9$^\dagger$ & 66.7$^\dagger$ & 59.4$^\dagger$ \\
\ssc{GPT-5-high}         &   50.0  &   30.8  &  15.9  & 67.1$^\dagger$ & 61.5$^\dagger$ & 58.0$^\dagger$ \\ \hline

\rowcolor{mygreen}\multicolumn{7}{c}{\textit{Open‑weight models}} \\ \hline
\ssc{Llama-3.2‑3B}                    &  1.3 & 0.0 & 0.0 &  5.4 &  2.0 & 0.0 \\
\ssc{Llama-3.1‑70B}                   &  6.0 & 2.0 & 1.8 &  8.7 &  6.1 & 0.0 \\
\ssc{Llama-4‑Scout-17B-16E (109B)}                   &  8.7 & 4.1 & 0.0 &  7.4 &  6.1 & 1.8 \\
\ssc{Qwen3-235B-A22B}                           &  6.7 & 2.0 & 0.0 & 12.8 & 16.3 & 3.6 \\
\ssc{DeepSeek‑R1‑Distill-Qwen‑1.5B}  &  0.7 & 2.0 & 1.8 &  2.7 &  0.0 & 0.0 \\
\ssc{DeepSeek‑R1‑Distill-Qwen‑14B}   & 10.7 & 6.1 & 0.0 & 11.4 & 14.3 & 5.4 \\
\deepseek                    & 35.6 & 8.2 & 0.0 & 14.8 &  6.1 & 5.4 \\
\ssc{\newdeepseek}                     & 27.5 & 10.2 & 5.4 & 19.5 & 14.3 & 5.4 \\
\ssc{gpt-oss-20b-high}                     & 3.4 & 0.0 & 2.9 & 8.9 & 11.4 & 6.3 \\
\ssc{gpt-oss-120b-high}                      & 14.4 & 2.6 & 7.2 & 23.3 & 7.7 & 8.7 \\
\bottomrule
\end{tabular}
\end{adjustbox}
\label{table:results_effective_year_full}
\end{table*}
\endgroup

\begingroup
\renewcommand{\arraystretch}{1.3}
\begin{table*}[ht!]
\centering
\caption{On \sealhard, performance degrades more when search results are uniformly unhelpful than when they contain conflicting answers.}
\begin{adjustbox}{max width=0.8\textwidth}
\begin{tabular}{lcccc}
\toprule
\multirow{2}{*}{\textbf{Model}} &
\multicolumn{2}{c}{\ssc{w/o search}} &
\multicolumn{2}{c}{\ssc{w/ search}} \\ \cline{2-5}
 & \ssc{unhelpful} & \ssc{conflicting} & \ssc{unhelpful} & \ssc{conflicting} \\ \hline

\rowcolor{myyellow}\multicolumn{5}{c}{\textit{Closed‑source models}} \\ \hline
\ssc{GPT-4o-mini}      &  7.2 & 10.4 & 10.9$^\dagger$ & 15.3$^\dagger$ \\
\ssc{GPT-4.1-mini}     & 10.0 & 16.6 & 10.0$^\dagger$ & 13.2$^\dagger$ \\
\ssc{GPT-4o}           &  9.0 & 13.8 & 11.8$^\dagger$ & 17.4$^\dagger$ \\
\ssc{GPT-4.1}          & 14.5 & 15.3 & 18.2$^\dagger$ & 22.2$^\dagger$ \\
\ssc{o3-mini-high}          & 10.9 & 13.9 & 8.2 & 13.9 \\
\ssc{o4-mini-high}    &   –  &   –  & 18.2$^\dagger$ & 20.1$^\dagger$ \\
\ssc{o3-high}         &   –  &   –  & 30.0$^\dagger$ & 34.7$^\dagger$ \\ 
\ssc{GPT-5-mini-high}         &   11.8  &   20.8  & 58.2$^\dagger$ & 61.8$^\dagger$ \\ 
\ssc{GPT-5-high}         &   36.4  &   38.9  & 62.7$^\dagger$ & 64.6$^\dagger$ \\ \hline

\rowcolor{mygreen}\multicolumn{5}{c}{\textit{Open‑weight models}} \\ \hline
\ssc{Llama-3.2‑3B}                    & 0.0 &  1.3 &  2.7 &  4.2 \\
\ssc{Llama-3.1‑70B}                   & 1.8 &  6.2 &  4.5 &  7.6 \\
\ssc{Llama-4‑Scout-17B-16E (109B)}                   & 3.6 &  7.6 &  4.5 &  6.9 \\
\ssc{Qwen3-235B-A22B}                           & 3.6 &  4.8 &  8.2 & 13.9 \\
\ssc{DeepSeek‑R1‑Distill-Qwen‑1.5B}  & 0.0 &  2.0 &  2.7 &  0.7 \\
\ssc{DeepSeek‑R1‑Distill-Qwen‑14B}   & 2.7 & 11.1 &  7.3 & 13.2 \\
\deepseek                    &20.9 & 23.6 &  9.1 & 12.5 \\
\ssc{\newdeepseek}                     & 18.2 & 20.1 & 11.8 & 18.1 \\
\ssc{gpt-oss-20b-high}                     & 1.8 & 3.5 & 5.4 & 9.7  \\
\ssc{gpt-oss-120b-high}                     & 6.4 & 13.9 & 7.3 & 24.3  \\

\bottomrule
\end{tabular}
\end{adjustbox}
\label{table:results_search_quality_full}
\end{table*}
\endgroup

\clearpage
\newpage
\section{\textsc{Seal-Hard} results by answer type}
\label{section:results_by_answer_type}
\begin{figure}[h!]
\centering
\includegraphics[width=\textwidth]{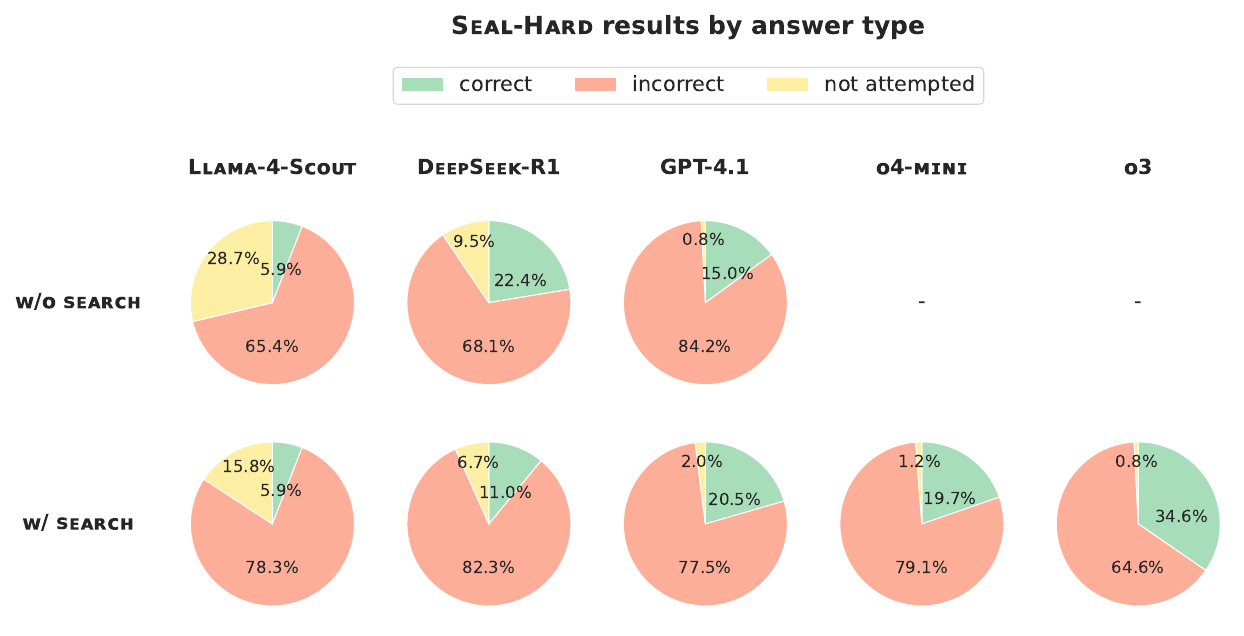}
\caption{On \sealhard, open-weight models like \ssc{Llama-4-Scout} and \ssc{DeepSeek-R1} choose to ``not attempt'' questions more often than proprietary models such as \ssc{GPT-4.1}, \ssc{o4-mini}, and \ssc{o3}.}
\label{figure:results_answer_type_fig}
\vspace{-2mm}
\end{figure}

Figure~\ref{figure:results_answer_type_fig} shows \sealhard results broken down by answer type: ``correct'', ``incorrect'', and ``not attempted''. We find that open-weight models like \ssc{Llama-4-Scout} and \ssc{DeepSeek-R1} choose to ``not attempt'' questions more often than proprietary models such as \ssc{GPT-4.1}, \ssc{o4-mini}, and \ssc{o3}. 

\newpage
\section{Human performance}
\begin{table}[h]
    \centering
    \caption{Performance of humans and frontier models on a subset of 50 \sealhard questions.}
    \begin{subtable}{0.4\linewidth}
    \centering
    \subcaption{Performance of frontier models}
    \begin{tabular}{lc}
    \toprule
       \textbf{Model}  & \textbf{Accuracy (\%)} \\
       \midrule
        \ssc{GPT-4o} & 6.0 \\
        \ssc{GPT-4.1} & 6.0 \\
        \ssc{o3-mini-high} & 8.0 \\
        \ssc{o4-mini-high} & 12.0 \\
        \ssc{o3-high} & 28.0 \\
        \bottomrule
    \end{tabular}
    \end{subtable}%
    \begin{subtable}{0.6\linewidth}
    \centering
    \subcaption{Performance of Humans}
        \begin{tabular}{lcc}
        \toprule
        \textbf{Overall accuracy (\%)} & \textbf{Open} & \textbf{Oracle} \\
        \midrule
         Average accuracy & 38.8 & 50.4 \\
         Best accuracy & 64.0 & 72.0 \\
         \midrule
         \textbf{Answer speed and accuracy} & & \\
         Share of answers given < 5 minutes & \multicolumn{2}{c}{52.8}  \\
         Accuracy of those fast answers & \multicolumn{2}{c}{53.0} \\
        \bottomrule
        \end{tabular}
    \end{subtable}
    \label{table:human_performance}
\vspace{-2mm}
\end{table}

\label{section:appendix_human_performance}

\vspace{-1mm}
\section{Qualitative analysis}
\label{section:appendix_qualitative_analysis}
Two authors independently evaluated 100 responses from six models: \ssc{GPT-4.1} (without search, with \ssc{FreshPrompt}, and with built-in search); \ssc{o3-mini}, \ssc{o3} (both under a medium reasoning effort); and \deepseek. Our analysis reveals clear differences across models in their reasoning and use of external knowledge. Among the \ssc{GPT-4.1} variants, the base model without search occasionally includes relevant \urls but often produces inaccurate answers due to outdated knowledge. The \ssc{FreshPrompt} version is better at detecting false-premise questions and tends to be more concise, though its accuracy depends heavily on retrieval quality. The built-in search variant produces more logically coherent answers and higher-quality citations, which supports factual verification, though it still exhibits occasional errors. We find that \ssc{o3} is capable of producing more informed and concise responses; however, it sometimes overthinks and mistakenly rejects valid answers. \ssc{o3-mini}'s outputs are easy to follow, yet the model occasionally misses relevant reasoning paths. Notably, both models generally acknowledge their knowledge cutoffs for time-sensitive queries, seek clarification, and suggest alternative strategies to support user decision-making. Finally, \deepseek tends to overthink and frequently repeats phrases like ``wait'', ``let me think'', and ``alternatively'' without arriving at a clear conclusion. Its lack of structured formatting also makes its responses harder to follow compared to \ssc{GPT-4.1} and \ssc{o3} models.

\clearpage
\section{Sample model outputs}
\small
\begin{longtblr}[
  caption = {Sample responses from several frontier models. Text in {\sethlcolor{lime}\hl{green}} shows when a model starts on the correct reasoning path but later shifts to an incorrect conclusion, which is shown in {\sethlcolor{pink}\hl{red}}.},
  label = {tab:examples},
]{
  colspec = {X[2]X[1.5]X[9.5]}, %
  rowhead = 1,
  hlines,
  row{even} = {myyellow},
}
\SetCell{c,m}\textbf{Question} & \SetCell{c,m}\textbf{Correct answer} &
\SetCell{c,m}\textbf{Model response} \\

\SetCell[c=3]{c} \ssc{o4-mini-high} \\ %

Who is the only female artist to have topped Spotify's list of most-streamed artists in three consecutive years (2020–2022)? 
& No one 
& **Recalling top artists**

Okay, let’s verify the most-streamed artists on Spotify by year. For 2020, it was Bad Bunny with 8.3 billion streams. In 2021, I initially thought Drake topped the list, and indeed, he did with about 9.2 billion, followed by Bad Bunny at 9.1 billion and J. Cole third. Moving to 2022, it looks like Bad Bunny returned to the top with 18 billion streams, confirming he was the most streamed artist again. I’ll need to check the latest articles for more accuracy.

**Identifying the top female artist**

Let’s figure out who the only female artist is that topped Spotify's list of most-streamed artists for three consecutive years, specifically from 2020 to 2022. Bad Bunny was the top male artist, so it can't be him. I think the question might refer to female artists, possibly indicating records within that category. 

Ariana Grande could be a candidate since she topped the female artist list in 2020. I need to confirm if she held that position in 2021 and 2022 as well. Let’s check!

**Clarifying the top female artist question**

I’m trying to break down whether the phrase “only female artist to have topped Spotify's list of most-streamed artists in three consecutive years” refers to a global category or just female artists. It seems like it might hint at the global list, but Bad Bunny and Drake held those spots in 2020 and 2021, {\sethlcolor{lime}\hl{suggesting no female artists topped that list globally}}. Instead, I think it might refer to female artist categories specifically, likely hinting at Taylor Swift, who seems to have topped the lists in those years—I'll need to confirm that!

**Confirming top female artist status**

Let’s clarify the question about which female artist topped Spotify’s list of most-streamed artists in three consecutive years from 2020 to 2022. I see that during this period, {\sethlcolor{lime}\hl{no female artist topped the global charts, as those positions went to male artists}}. It seems the question likely refers to the female category specifically. Given that Taylor Swift has been highlighted as the top female artist for those years, it appears she's the answer. I'll double-check the details to ensure accuracy!

**Identifying the top artist**

I've figured it out: {\sethlcolor{pink}\hl{the answer is Taylor Swift. From 2020 to 2022, she was Spotify's most-streamed female artist globally. This makes her the only female artist to top that list for three consecutive years}}. Although there might be some confusion with Ariana Grande, it's definitely Taylor who holds this title. If I summarize my findings, I can simply state that Taylor Swift was the most-streamed artist during those years, confirming my final answer clearly as Taylor Swift.

\textbf{Final answer}: Taylor Swift. She was Spotify’s most-streamed (female) artist globally in 2020, 2021 and 2022—making her the only woman to top that chart three years running.
 \\

\pagebreak
\SetCell[c=3]{c} \ssc{o3-high} \\

How many times has a U.S. team won the ICPC World Finals after 1997—a period largely dominated by Russian and Chinese teams? 
& One 
& **Counting ICPC Wins for U.S. Teams**

I'm analyzing how many times a U.S. team has won the ICPC World Finals since 1997, a period dominated by Russian and Chinese teams. After verifying the winners from 1998 to the present (2024), it's clear that I need to count the instances where the winning university is from the U.S. Based on my recap, it seems like there might be two wins. My next step is to confirm this with the full list of winners.

**Recalling ICPC Winners**

I’m compiling the winners of the ICPC from the years following 1997. The last U.S. win I note is from 1997, when Harvey Mudd College triumphed. After that, the champions seem mostly to come from Russia and China. I’m looking back at the details year by year to see if there were any U.S. winners in the years following 1997. So far, {\sethlcolor{pink}\hl{it looks like there might not be many left}}, which I’ll confirm soon.

**Confirming U.S. ICPC Wins**

I'm sorting through the ICPC winners since 1998 to determine how many times a U.S. team has won since then, starting with the list of champions. It looks like after checking the winners, St. Petersburg ITMO and Moscow State University frequently took the title. However, {\sethlcolor{lime}\hl{I want to confirm that there was indeed a U.S. win in 2022 by MIT. Overall, I suspect there's only one U.S. win after 1997}}, which I’ll clarify soon.

\textbf{Final answer}: {\sethlcolor{pink}\hl{Zero. Since Harvey Mudd College’s victory in 1997, no U.S. university has won the ICPC World Finals}}.
 \\

\SetCell[c=3]{c} \ssc{GPT-5-high} \\
Which two countries share the longest continuous international border in the world by total length?
& Kazakhstan and Russia
& **Answering border question**

I need to answer which two countries share the longest continuous international border. That would be Canada and the United States, which is about 8,891 kilometers (or 5,525 miles) long, including Alaska. I want to be clear, as it’s often quoted this way. {\sethlcolor{pink}\hl{The question highlights that it's a stable fact, so I don’t need to look anything up}}. Keeping it concise is essential, but mentioning Alaska might be good for clarity.
**Clarifying the border definition**

I need to be cautious about using the web.run tool since the fact I’m addressing is stable and won’t change—it’s not necessary to browse. For a concise response, I’ll state that Canada and the United States share the longest continuous international border, approximately 8,891 kilometers (or 5,525 miles). The term ""continuous"" might imply that it's not separated by maritime boundaries. While it’s punctuated by water in the Great Lakes, it's still considered continuous because those areas are recognized as part of the international border.

\textbf{Final answer}: Canada and the United States (about 8,891 km / 5,525 miles).
 \\

\pagebreak
\SetCell[c=3]{c} \ssc{DeepSeek-R1} \\
Which tropical rainforest represents a quarter of the total area of remaining tropical forests on Earth and comprises one of the largest and most biodiverse tracts of tropical rainforest in the world?
& Congolian rainforests
& Okay, so the question is asking which tropical rainforest represents a quarter of the remaining tropical forests and is one of the largest and most biodiverse. Let me think.

First, I know the major tropical rainforests are the Amazon, Congo Basin, and the Southeast Asian ones like Borneo and New Guinea. The Amazon is the biggest, right? But I need to confirm if it's a quarter of the remaining area. The Congo Basin is also huge. Then there's the issue of biodiversity. The Amazon is known for its biodiversity, but so are some others. 

Wait, the question says ""a quarter of the total area of remaining tropical forests."" So which one is that? The Amazon is often cited as the largest, covering around 5.5 million square kilometers. If the total remaining tropical forests are, say, around 20 million square kilometers, then the Amazon would be about a quarter. Let me check approximate numbers. I remember that the total tropical forest area is decreasing, but estimates might vary. Some sources say {\sethlcolor{lime}\hl{the Amazon is about half of the world's remaining rainforests}}. Wait, that might not be right. Maybe the Amazon is about a third? Hmm. 

Alternatively, maybe the Congo Basin is second, but the Amazon is definitely larger. If the total remaining tropical rainforest is, say, 10 million square km, then the Amazon being 5.5 million would be more than half. But maybe the total is higher. Let me think. According to some sources, tropical rainforests cover about 17-20 million square km globally. {\sethlcolor{pink}\hl{The Amazon is around 5.5 million, so that's roughly a third}}. Wait, but the question says a quarter. Maybe the answer is the Amazon, but the percentage is a quarter. Maybe the total remaining is higher. Alternatively, perhaps the user is referring to the Amazon as a quarter. 

Also, the part about ""comprises one of the largest and most biodiverse tracts."" The Amazon is definitely one of the most biodiverse. The Congo Basin is also biodiverse but maybe less so. Southeast Asian rainforests are biodiverse too but smaller in area. So putting it all together, the answer is likely the Amazon Rainforest. But I need to make sure about the percentage. If the total remaining tropical rainforests are approximately 20 million km², then {\sethlcolor{pink}\hl{the Amazon at 5.5 million would be about 27.5\%, which is roughly a quarter}}. So that fits. Therefore, the answer is the Amazon Rainforest.

\textbf{Final answer}: The **Amazon Rainforest** represents approximately a quarter of the total area of remaining tropical forests on Earth and is one of the largest and most biodiverse tracts of tropical rainforest in the world. 

- **Size**: Spanning around 5.5 million square kilometers (2.1 million square miles), it covers about 27\% of the globe's remaining tropical forest area (estimated at ~20 million km² total).  

- **Biodiversity**: It hosts 10\% of known species, including iconic wildlife like jaguars, pink river dolphins, and countless plant and insect species.  

- **Geographic Reach**: It spans nine countries, with the majority in Brazil, and plays a critical role in global carbon storage and climate regulation.  

This immense ecosystem is often termed the ""lungs of the Earth"" due to its vital role in absorbing carbon dioxide and producing oxygen.
 \\
\end{longtblr}

\label{section:appendix_sample_model_outputs}

\clearpage
\section{Additional \textsc{Seal-0} results}
\begin{table*}[h]
\centering
\caption*{$^\dagger$ indicates results using the model's built-in search; all other search-based results use \ssc{FreshPrompt}~\citep{vu-etal-2024-freshllms}.}
\caption{\sealzero poses a significant challenge for a  range of both closed and open-source models.}

\begin{tabular}{l c}
\toprule
\textbf{Model} & \textbf{Accuracy} \\
\hline

\rowcolor{myyellow}\multicolumn{2}{c}{\textit{Closed-source models}} \\ \hline
\ssc{Grok 4}                & 20.7$^\dagger$ \\
\ssc{Gemini 2.5 Pro}        & 19.8$^\dagger$ \\
\ssc{o3-pro}                & 18.9$^\dagger$ \\
\ssc{o3}                    & 15.3$^\dagger$ \\
\ssc{Gemini 2.5 Flash}      & 13.5$^\dagger$ \\
\ssc{o4-mini}               & 6.3$^\dagger$ \\
\ssc{Grok 3}                & 5.4$^\dagger$ \\
\ssc{Gemini 2.5 Flash-Lite} & 2.7$^\dagger$ \\
\ssc{o3-mini}               & 2.7 \\
\ssc{Grok 3 Mini}           & 2.7$^\dagger$ \\
\ssc{GPT-4.1}               & 0.0$^\dagger$ \\
\ssc{GPT-4.1 mini}          & 0.0$^\dagger$ \\ \hline

\rowcolor{mygreen}\multicolumn{2}{c}{\textit{Open‑weight models}} \\ \hline

\ssc{Qwen3-235B-A22B}                 & 5.4 \\
\ssc{DeepSeek-R1-671B}           & 4.5 \\
\ssc{Llama-4-Scout-17B-16E (109B)}         & 0.0 \\
\bottomrule
\end{tabular}
\label{table:additional_seal0_results}
\end{table*}

\label{section:additional_seal0_results}

\end{document}